\documentclass[lettersize,journal]{IEEEtran}

\usepackage{amsmath,amsfonts}

\usepackage{array}
\usepackage{multirow, array, ragged2e, lscape, eqparbox, xspace, setspace}
\usepackage{booktabs, tabularx,tabulary}

\usepackage[caption=false,font=normalsize,labelfont=sf,textfont=sf]{subfig}
\usepackage{graphicx}

\usepackage[colorlinks = true,
            linkcolor = blue,
            urlcolor  = purple,
            citecolor = blue,
            anchorcolor = blue]{hyperref}
\usepackage{url}
\usepackage{cite}
\usepackage{orcidlink}

\usepackage{textcomp}
\usepackage{longtable}
\usepackage{stfloats}
\usepackage{verbatim}
\usepackage{makecell}
\usepackage{pifont}
\usepackage{enumitem}  
\usepackage{textcomp}
\usepackage{forest}
\usepackage{enumitem}

\makeatletter
\newcolumntype{P}[1]{>{\@minipagetrue}p{#1}}
\makeatother

\graphicspath{{figures/}}

\usepackage{hyphenat}

\def\mybar#1{
  #1 & {\color{gray}\rule{#1cm}{8pt}}}

\makeatletter
\DeclareRobustCommand\onedot{\futurelet\@let@token\@onedot}
\def\@onedot{\ifx\@let@token.\else.\null\fi\xspace}

\makeatother

\begin{document}

\title{ A New Era in Computational Pathology: A Survey on Foundation and Vision-Language Models}

\author{Dibaloke Chanda \orcidlink{0000-0001-5993-659X},~\IEEEmembership{Graduate Student Member,~IEEE}, \\ Milan Aryal \orcidlink{0009-0005-1326-9804},\\ Nasim Yahya Soltani \orcidlink{0000-0002-4502-8715},~\IEEEmembership{Member,~IEEE} \\  
Masoud Ganji,
\IEEEmembership{Fellow,~CAP}
\thanks{}
\thanks{}}

\markboth{}%
{Chanda \MakeLowercase{\textit{et al.}}: A New Era in Computational Pathology: A Survey on Foundation and Vision-Language Models}


\maketitle

\begin{abstract}
Recent advances in deep learning have completely transformed the domain of computational pathology~(CPath). More specifically, it has altered the diagnostic workflow of pathologists by integrating foundation models (FMs) and vision-language models (VLMs) in their assessment and decision-making process. The limitations of existing deep learning approaches in CPath can be overcome by FMs through learning a representation space that can be adapted to a wide variety of downstream tasks without explicit supervision. Deploying VLMs allow pathology reports written in natural language be used as rich semantic information sources to improve existing models as well as generate predictions in natural language form. In this survey, a holistic and systematic overview of recent innovations in FMs and VLMs in CPath is presented. Furthermore, the tools, datasets and training schemes for these models are summarized in addition to categorizing them into distinct groups. This extensive survey highlights the current trends in CPath and its possible revolution through the use of FMs and VLMs in the future.

\end{abstract}

\begin{IEEEkeywords}
Computational pathology, foundation models, multi-modal, vision-language models.
\end{IEEEkeywords}

\section{Introduction}\label{sec:Introduction}
\IEEEPARstart{I}{\lowercase{n}} recent years there has been a surge of artificial intelligence (AI)-based approaches~\cite{srinidhi2021deep,morales2021artificial,echle2021deep,van2021deep,abdelsamea2022survey,kim2022application,shmatko2022artificial,waqas2023revolutionizing,asif2023unleashing,atabansi2023survey,cooper2023machine,song2023artificial,xu2023vision,bahadir2024artificial,mcgenity2024artificial,hosseini2024computational,brussee2024graph,gadermayr2024multiple,cheng2024applications} in computational pathology (CPath) owing to wide adoption of digital slide scanners. In fact, large-scale curation and annotation~\cite{weinstein2013cancer,kang2021development} of whole slide images (WSIs) has ensured adequate data for training of these AI-based models. The goal of generating such AI-based models is to automate and expedite the diagnosis and prognosis process of CPath. Traditional diagnosis process in CPath is time-consuming and requires experts with extensive domain knowledge. In addition, the wide range of cancer types and grades as well as the heterogeneity among diagnosis/prognosis tasks, makes it challenging to come up with a unified general approach. 

Several research studies have addressed the issue with a unified approach and among the proposed methods, foundation models (FMs) have gained a lot of attention in recent years\cite{zimmermann2024virchow,hua2024pathoduet,vorontsov2024foundation,hoptimus0,dippel2024rudolfv,nechaev2024hibou,yang2024foundation,juyal2024pluto,zhou2024knowledge,chen2024towards,campanella2024computational,filiot2023scaling,kang2023benchmarking,wang2022transformer,ma2024towards,lu2024multimodal,xu2024whole,shaikovski2024prism,sun2024pathasst,lu2024visual,huang2023visual}. More specifically, FMs leverage self-supervised learning (SSL)~\cite{gui2024survey} schemes to learn a rich representation in a task-agnostic manner. Owing to self-supervised pre-training (SSPT), FMs do not require large-scale annotated data which is hard to come by in CPath. Furthermore, these models can be trained with a diverse selection of datasets containing tissue samples from different organs and associated with different cancer types, as well as scanner types. Certain research works focus on making sure this diversity is present in the pre-training dataset~\cite{dippel2024rudolfv,juyal2024pluto}. As a result, the resultant pre-trained model can easily be utilized in a wide range of downstream tasks while maintaining robustness to extreme variation in tissue samples.

\begin{figure}[!htt]
    \centering
    \includegraphics[width=0.55\linewidth]{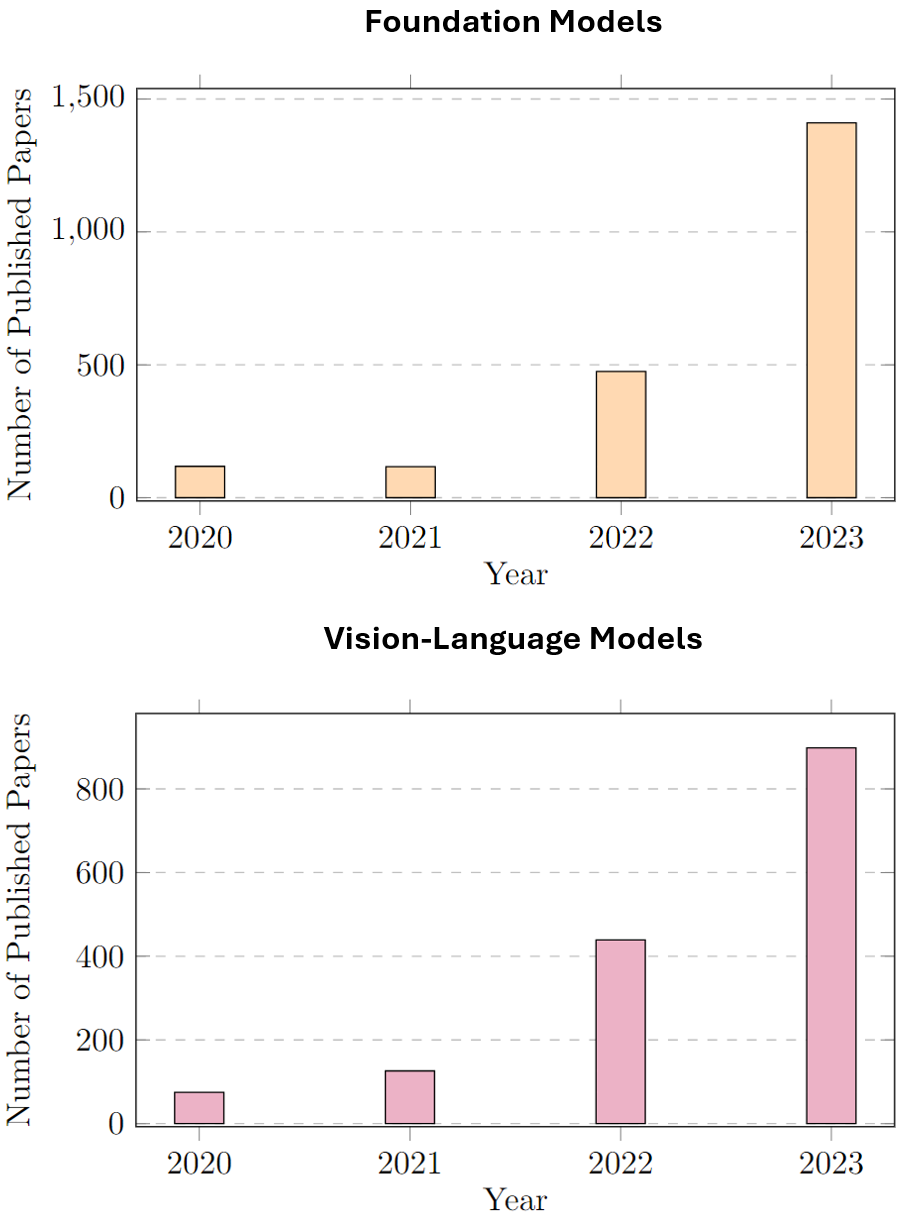}
    \caption{Number of publications for FMs and VLMs in pathology (from Google Scholar). The search keywords include ``vision-language" + ``pathology" for VLMs statistics and ``foundation models"+ ``pathology" for FMs statistics.}
    \label{fig:publication_stat}
\end{figure}

\begin{figure*}[t]
    \centering
    \includegraphics[width=0.74\linewidth]{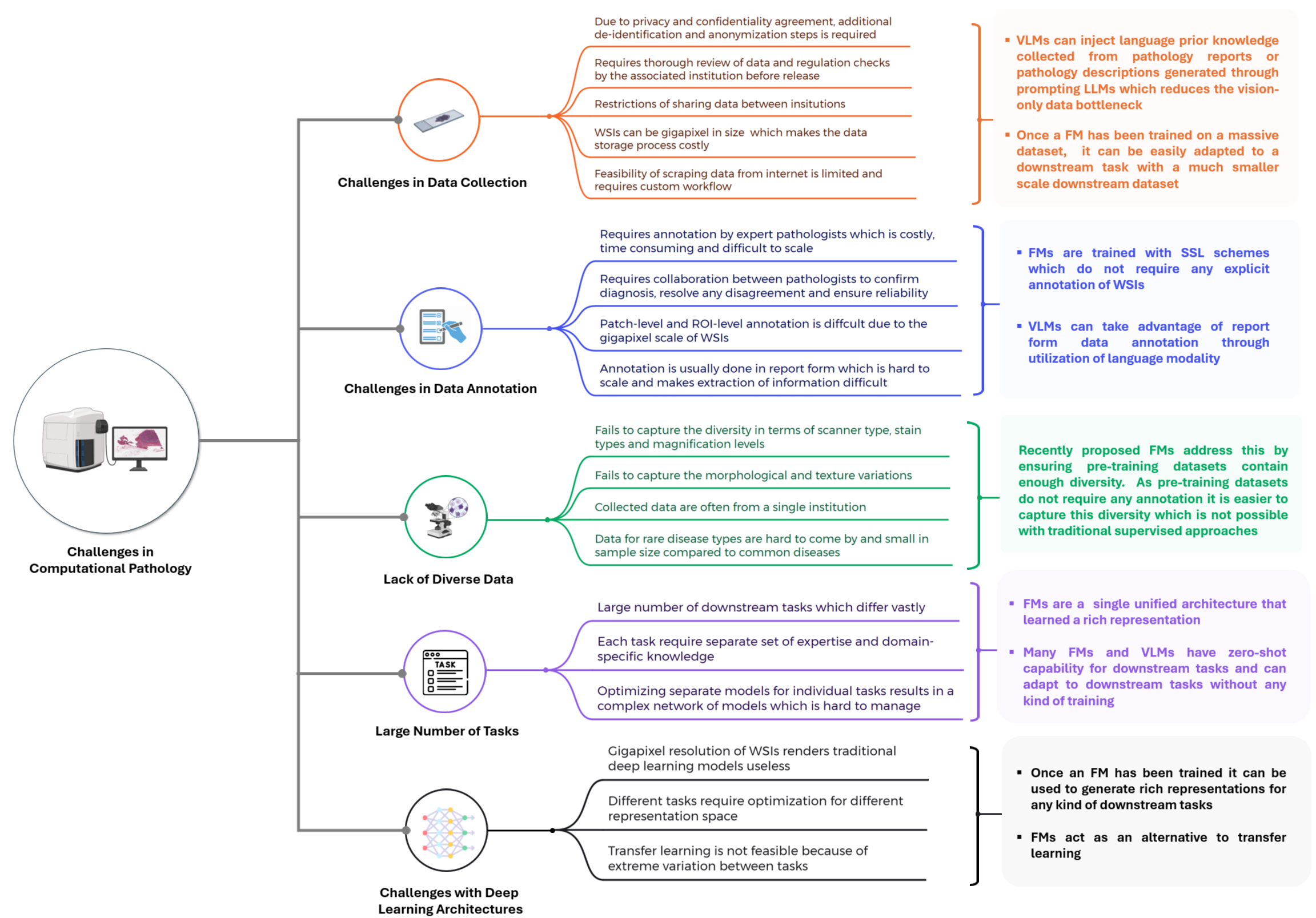}
    \caption{Outline of major challenges in CPath. Several causes and consequences for each challenge are outlined in addition to how FMs and VLMs address these challenges.}
    \label{fig:five_challenges}
\end{figure*}

The impact of FMs in CPath can be amplified by integrating the power of vision-language models (VLMs) ~\cite{zhang2024vision}. More specifically, VLMs have surged in popularity after the introduction of contrastive language-image pre-training (CLIP) model by OpenAI~\cite{radford2021learning}. Pathology reports, books, educational videos, etc. are rich sources of semantic information that can be utilized by VLMs to significantly boost the performance. This is not typically possible with vision-only models. When deployed in conjunction with FMs, they can perform as AI-based pathologists capable of performing a vast array of tasks as evidenced by recent literature~\cite{lu2024multimodal}. {Even though CPath is a specialized area of study in the medical field, there is a large increase in the number of publications focusing on FMs and VLMs in pathology as shown in Fig.~\ref{fig:publication_stat} demonstrating a dominant future direction of this field.}

To appreciate the full impact of FMs and VLMs in CPath, the major challenges in CPath are outlined in Fig.~\ref{fig:five_challenges}. In addition, the way FMs and VLMs address these challenges is also mentioned.

\subsection{Scope of the Review}
In this review, the main emphasis is put on the application of FMs and VLMs in CPath, especially the details of their architectures and training schemes. Note that these two categories are not mutually exclusive, meaning that some research articles belong to both categories which are vision-language foundation models (VLFMs). In addition, details of multi-modal datasets are summarized with a focus on vision and language as the modalities.
\begin{table}[!htt]
\caption{Number of Surveyed articles published in top journals and conferences within 2021-2024}
\centering
\renewcommand{\arraystretch}{1.6}
\setlength{\tabcolsep}{3pt}
\resizebox{0.8\linewidth}{!}{
\begin{tabular}{l|cl}
\bottomrule
\textbf{Journal/Conference Venue} & \multicolumn{2}{l}{\textbf{Counts}} \\ \hline
    
    Nature/Nature Medicine            &    \mybar{6}   \\

    CVPR              &    \mybar{8}     \\
    NeurIPS            &    \mybar{2}     \\

    ECCV/WACV/ICCV        &   \mybar{5}      \\
       MICCAI            &   \mybar{5}      \\
          AAAI            &   \mybar{2}      \\

   Elsevier/Springer            &   \mybar{6}      \\
\toprule

\end{tabular}}
\label{tab:venue_stat}
\end{table}
Several self-imposed rules and restrictions were used as guidelines throughout the review to ensure the scope of the paper is maintained. 
\begin{enumerate}
    \item  First, articles that focus solely on pathology are included. So, and articles with a focus on other areas of the biomedical domain are excluded. As an example, articles like BiomedCLIP~\cite{zhang2023biomedclip} with pathology as a subsection of the research are not included in this review paper.

    \item Secondly, only vision-language models are included and articles that use other modalities along with vision are excluded. As an example, along with vision, transcriptomics\cite{jaume2024modeling} can be used to address pathology tasks. However, such papers are excluded to maintain the scope.

    \item  Both peer-reviewed articles and pre-print articles are included in the survey. Among the peer-reviewed articles, a significant number of articles are published in top-tier journals and conferences as shown in Table~\ref{tab:venue_stat}.
    
\end{enumerate}
\begin{figure*}
    \centering
    \includegraphics[width=\linewidth]{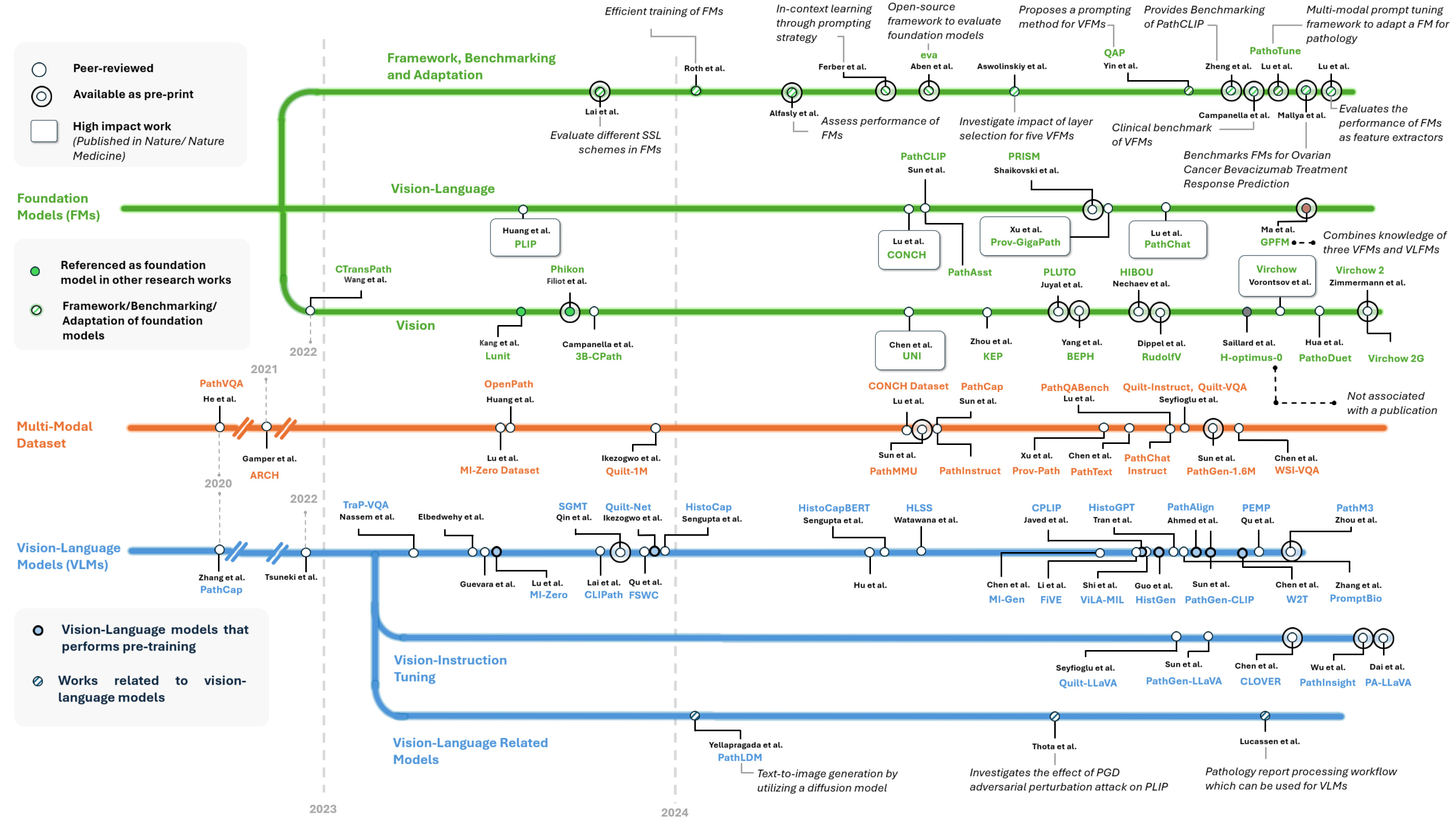}
    \caption{Visualization of the timeline of recently published work in CPath utilizing FMs and VLMs as well as multi-modal datasets. To maintain transparency, we clearly annotate research articles that have been peer-reviewed and articles that are available as pre-prints. Furthermore, high-impact pioneering research works published in prominent journals are highlighted. For pre-prints if there are multiple versions, the latest version and the corresponding date are used.}
    \label{fig:timeline}
\end{figure*}
\subsection{Contribution and Organization}
In the past few years, quite a few review articles~\cite{srinidhi2021deep,morales2021artificial,echle2021deep,van2021deep,abdelsamea2022survey,kim2022application,shmatko2022artificial,waqas2023revolutionizing,asif2023unleashing,atabansi2023survey,cooper2023machine,song2023artificial,xu2023vision,bahadir2024artificial,mcgenity2024artificial,hosseini2024computational,brussee2024graph,gadermayr2024multiple,cheng2024applications} have been published focusing on CPath. Most of these articles provide a review of the application of deep learning as a whole rather than focusing on a specific subtopic. There are handful of papers that focus on specific architecture like MIL~\cite{gadermayr2024multiple}, graph-based models~\cite{ahmedt2022survey,brussee2024graph}, transformer-based models~\cite{xu2023vision,atabansi2023survey}, LLMs~\cite{cheng2024applications}, etc. The contributions of this review article are summarized below: 

\begin{enumerate}
    \item To the best of our knowledge, this is the first review article to summarize recent advances (most articles from 2023-2024) in FMs and VLMs (Section \ref{sec:foundation_models} and \ref{sec:vision_language_models}). The closest peer-reviewed survey article~\cite{waqas2023revolutionizing} mostly provides a high-level overview without going into the details of FMs.  

    \item An exhaustive list of multi-modal datasets (Section  \ref{sec:multi_modal_datasets}) in CPath that are being used or can be used in vision-language research are outlined.

    \item Given the diverse datasets and architectures, descriptions for individual research are provided in tabular format~(Table~\ref{tab:multi-modal_datasets}, Table~\ref{tab:vision_pre_training}, Table~\ref{tab:vision_language_pretraing}, Table~\ref{tab:vlms}) so it is easier for readers to follow. 

    \item Annotated timeline of the surveyed articles (Fig.~\ref{fig:timeline}) which provides a clear idea of the evolution of the FMs and VLMs in CPath. In addition, a project page is available (\href{https://cpath-fms-vlms.github.io/survey/}{https://cpath-fms-vlms.github.io/survey/}) with additional information about the surveyed articles.
\end{enumerate}

The rest of the article is organized as follows: in section~\ref{sec:multi_modal_datasets} existing multi-modal datasets are listed along with details about their source, pre-processing techniques, etc. Section~\ref{sec:foundation_models} outlines existing FMs in CPath and descriptions for the vision and vision-language pre-training schemes for those FMs are provided. In section~\ref{sec:vision_language_models} an extensive list of VLMs in CPath is presented along with details of their architectures, utilized datasets and contributions. Finally, section~\ref{sec:conclusion} concludes the paper.

\section{Multi-Modal Datasets in Pathology}\label{sec:multi_modal_datasets}

In this section, a comprehensive summary of the existing multi-modal datasets available in CPath is provided. The modalities taken into account are vision and language. All key information regarding each dataset is summarized in Table~\ref{tab:multi-modal_datasets}.

\begin{figure}[!htt]
    \centering
\includegraphics[width=0.55\linewidth]{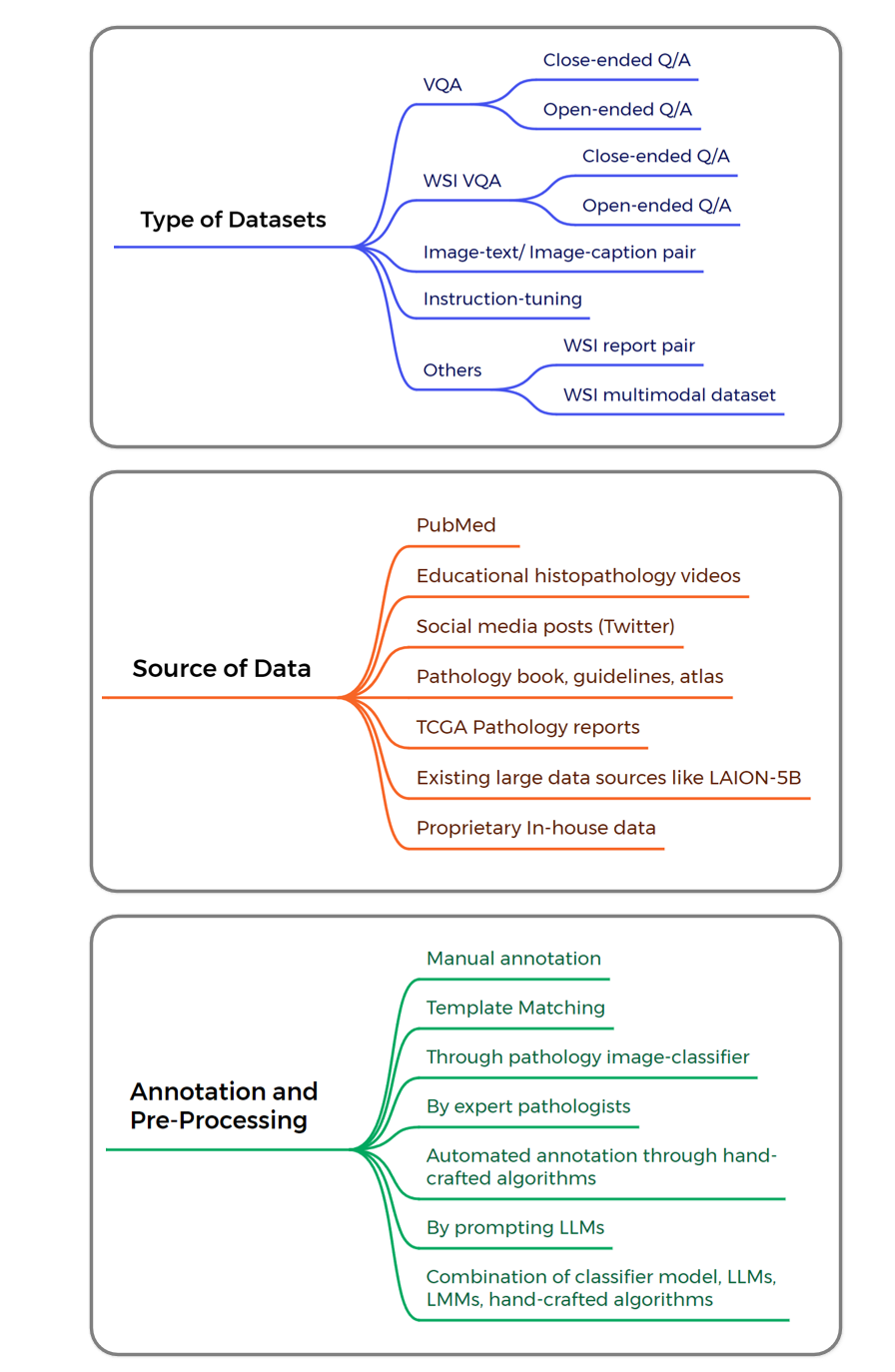}
    \caption{Different components of multi-modal datasets in computational pathology: Type of datasets, sources of data, annotation and pre-processing}
    \label{fig:multi-modal_aspects}
\end{figure}

To have a comprehensive understanding of the existing multi-modal datasets for pathology, three different components need to be considered as shown in Fig.~\ref{fig:multi-modal_aspects}.

\begin{table*}[p]
\caption{Summary of Multi-Modal Datasets in Pathology}
\centering
\renewcommand{\arraystretch}{1.5}
\setlength{\tabcolsep}{4pt}
\resizebox{0.98\linewidth}{!}{
\begin{tabular}{lcllcllcl}
\toprule
\multicolumn{1}{l}{\textbf{Dataset}} &
  \multicolumn{1}{c}{\textbf{\makecell[c]{Type of \\ Data}}} &
  \multicolumn{1}{c}{\textbf{Size}} &
  \multicolumn{1}{c}{\textbf{\makecell[c]{Image and \\ Text Source}}} &
  \textbf{\makecell[c]{Method of \\ Generation/Procurement}} &
  \multicolumn{1}{l}{\textbf{\makecell[l]{Other Utilized Dataset/Models \\ for Generation/Procurement}}} &
  \multicolumn{1}{c}{\textbf{\makecell[c]{Dataset \\ Used By}}} &
  \textbf{\makecell[c]{Variation/ \\ Subset/Extension}} &
  \multicolumn{1}{c}{ \makecell[c]{\textbf{Availability} (\textit{Linked})} } \\\bottomrule

WSI-VQA~\cite{chen2024wsi} &
  \makecell[c]{WSI and \\Q/A pairs} &
  \makecell[c]{977 WSIs \\ 8,672 question-answer \\ pairs (slide-level) \\ with average 8.9 \\ Q/A pairs per WSI} &
  \makecell[c]{WSI and pathology\\ report from \\ TCGA-BRCA}&
  \makecell[c]{Prompting LLMs \\ and template \\ matching heuristics }& 
  \makecell[l]{\textbf{Model/Framework:}\\ GPT-4~\cite{achiam2023gpt}} &
  \makecell[c]{Wsi2Text\\ Transformer \\ (W2T)~\cite{chen2024wsi}} &
  \makecell[c]{4535 \\ close-ended \\ VQA subset and \\ 4137 \\ open-ended \\ VQA subset} & \makecell[c]{\href{https://github.com/cpystan/WSI-VQA}{\ding{51}}}
   \\
  \midrule

  PathGen-1.6M~\cite{sun2024pathgen} &
  \makecell[c]{Image-caption \\ pairs} &
  \makecell[c]{1.6 million} &
  \makecell[c]{WSI and pathology\\ report from TCGA }&
  \makecell[c]{Multi-agent \\ collaboration \\ and caption generation \\ with large\\ multimodal models\\(LMMs) }& 
  \makecell[l]{\textbf{Dataset:}\\ PathCap~\cite{sun2024pathasst}, Quilt-1M~\cite{ikezogwo2024quilt}, \\ OpenPath~\cite{huang2023visual} \vspace{0.3em} \\  \textbf{Model/Framework:} \\  LLaVA-v1.5~\cite{liu2024visual}, Vicuna~\cite{vicuna2023} \\ OpenCLIP~\cite{ilharco_gabriel_2021_5143773}, GPT-4~\cite{achiam2023gpt}} &
  \makecell[c]{PathGen-CLIP\cite{sun2024pathgen}, \\ PathGen-LLaVA~\cite{sun2024pathgen}} &
  \makecell[c]{200K \\ instruction-tuning \\ data (Extension) } & \makecell[c]{\href{https://github.com/superjamessyx/PathGen-1.6M}{\ding{51}}}
   \\
   \midrule

Quilt-Instruct~\cite{seyfioglu2024quilt} &
  \makecell[c]{Instruction-tuning \\ question-answer \\ pairs} &
  \makecell[c]{107,131 \\ histopathology-specific\\ Q/A pairs } &
  \makecell[c]{Over 1,000 hours of \\ 4,149 educational \\ histopathology \\ videos
from \\ YouTube}&
  \makecell[c]{Prompting GPT-4, \\ hand-crafted algorithms \\
  for extraction of \\video frames and \\ spatial annotation}& 
  \makecell[l]{\textbf{Dataset:}\\ Quilt-1M~\cite{ikezogwo2024quilt}\vspace{0.3em} \\  \textbf{Model/Framework:} \\ GPT-4~\cite{achiam2023gpt}} &
  \makecell[c]{Quilt-LLaVA\cite{seyfioglu2024quilt}} &
  \makecell[c]{\textbf{Quilt-VQA}\\ with 985 images \\ 1,283 Q/A pairs \\ where 940 are\\ open-set and \\
  343 closed-set
  } & \makecell[c]{\href{https://github.com/aldraus/quilt-llava}{\ding{51}}}
   \\
   \midrule

   PathQABench~\cite{lu2024multimodal} &
  \makecell[c]{ROI-annotated WSI \\ and Q/A} &
  \makecell[c]{48 H\&E WSIs \\ (25 WSIs from \\PathQABench-Private\\  23 WSIs from \\ PathQABench-Public)\\ + 48 close-ended Q/A \\ 115 open-ended Q/A} &
  \makecell[c]{PathQABench-Private\\ from \\ private in-house cases \\PathQABench-Public\\ from \\ TCGA cases }&
  \makecell[c]{Expert-pathologists \\ curated and \\ annotated}& 
  \makecell[l]{\textbf{---}} &
  \makecell[c]{PathChat\cite{lu2024multimodal}} &
  \makecell[c]{(Subsets) \\ \textbf{PathQABench-Public}\\ and \\ \textbf{PathQABench-Private} \\ \\  \textbf{PathChatInstruct}  \\ 456,916\\ instruction-tuning \\  dataset (Extension)
  } & \makecell[c]{\\ \href{https://github.com/fedshyvana/pathology_mllm_training}{\ding{51}}}
   \\
   \midrule

PathText~\cite{chen2023mi} &
  \makecell[c]{WSI and \\question-text \\ pairs} &
  \makecell[c]{1,041 WSI \\9,009 WSI-text\\ pairs} &
  \makecell[c]{WSI and pathology\\ report from \\ TCGA-BRCA}&
  \makecell[c]{Prompting LLMs, \\ OCR, manual \\ annotation, classifier}& 
  \makecell[l]{\textbf{---}} &
  \makecell[c]{MI-Gen~\cite{chen2023mi}} &
  \makecell[c]{\textbf{---}} & \makecell[c]{ \href{https://github.com/cpystan/Wsi-Caption}{\ding{51}}}
   \\
  \midrule

        Prov-Path~\cite{xu2024whole} &
  \makecell[c]{WSI\\ and reports pairs} &
  \makecell[c]{17,383 \\ WSI-reports pairs} &
  \makecell[c]{ Proprietary dataset \\ from Providence\\ 
Health System (PHS)}&
  \makecell[c]{K-means to\\ generate four \\ representative\\ reports which are \\ used to prompt \\ GPT 3.5 to clean \\ rest of the reports
}& 
  \makecell[l]{\textbf{Model/Framework:} \\ 
  GPT-3.5} &
  \makecell[c]{Prov-GigaPath~\cite{xu2024whole}} &
  \makecell[c]{\textbf{---}} & \makecell[c]{\href{https://github.com/prov-gigapath/prov-gigapath}{\ding{51}}}
   \\
   \midrule

      PathCap~\cite{sun2024pathasst} &
  \makecell[c]{Image-caption \\ pairs} &
  \makecell[c]{207K} &
  \makecell[c]{PubMed, \\internal pathology\\
guidelines books, \\ annotation by \\expert cytologists}&
  \makecell[c]{Parsing from PubMed, \\ image processing \\ with 
  YOLOv7,\\  ConvNeXt, PLIP, \\ caption refinement \\ and text processing \\ with ChatGPT}& 
  \makecell[l]{\textbf{Dataset:}\\ PubMed\vspace{0.3em} \\  \textbf{Model/Framework:} \\ YOLOv7~\cite{wang2023yolov7}, ConvNeXt~\cite{liu2022convnet}\\ ChatGPT~\cite{openai2023chatgpt}, PLIP~\cite{huang2023visual}} &
  \makecell[c]{PathAsst~\cite{sun2024pathasst}} &
  \makecell[c]{\textbf{PathInstruct} \\ 180K pathology \\ multimodal \\
instruction-following \\ samples (Extension)
  } & \makecell[c]{\href{https://github.com/superjamessyx/Generative-Foundation-AI-Assistant-for-Pathology}{\ding{51}}}
   \\
      \midrule

PathMMU~\cite{sun2024pathmmu} &
  \makecell[c]{Image and \\Q/A \\ pairs} &
  \makecell[c]{24,067 pathology images \\ 33,428 Q/A } &
  \makecell[c]{PubMed, pathology atlas,  \\ educational videos,\\pathologist-shared \\ images with explanations\\ on Twitter}&
  \makecell[c]{Prompting GPT-4, \\ heuristics, \\ annotation by \\ seven pathologists}& 
  \makecell[l]{\textbf{Dataset:}\\ Quilt-1M~\cite{ikezogwo2024quilt}, OpenPath~\cite{huang2023visual} \vspace{0.3em} \\  \textbf{Model/Framework:} \\ GPT-4~\cite{achiam2023gpt}, YOLOv6~\cite{li2022yolov6}} &
  \makecell[c]{\textbf{---}} &
  \makecell[c]{PubMed, SocialPath\\EduContent, Atlas, \\PathCLS (Subsets)} & \makecell[c]{\href{https://github.com/PathMMU-Benchmark/PathMMU}{\ding{51}}}
   \\
   \midrule

         \makecell[l]{Dataset from \\CONCH ~\cite{lu2024visual}} &
  \makecell[c]{Image-caption \\ pairs} &
  \makecell[c]{1,786,362 \\ image-caption pairs} &
  \makecell[c]{PubMed, \\publicly available \\
research articles,\\ internal data from \\ Mass General \\Brigham institution}&
  \makecell[c]{Hand-crafted \\ workflow \\ with YOLOv5, \\ CLIP, 
BioGPT
}& 
  \makecell[l]{\textbf{Model/Framework:} \\ 
  YOLOv5~\cite{redmon2016you}, CLIP\\BioGPT~\cite{luo2022biogpt}} &
  \makecell[c]{CONCH~\cite{lu2024visual}} &
  \makecell[c]{\textbf{PMC-Path} \\ (data from   PubMed)  \\ \textbf{EDU} \\(data extracted from\\ educational notes)} & \makecell[c]{\href{https://github.com/mahmoodlab/CONCH}{\ding{51}}}
   \\
      \midrule

QUILT~\cite{ikezogwo2024quilt} &
  \makecell[c]{Image-text \\ pairs} &
  \makecell[c]{437,878 images \\ aligned with \\802,144 \\
text pairs} &
  \makecell[c]{1,087 hours of 4,504\\ narrative  educational \\ histopathology \\ videos
from \\ YouTube}&
  \makecell[c]{Prompting LLMs, \\ hand-crafted algorithms,\\ human knowledge \\ databases (UMLS),\\
automatic speech \\ recognition}& 
  \makecell[l]{\textbf{Dataset:}\\ OpenPath~\cite{huang2023visual}, PubMed, \\ LAION-5B~\cite{schuhmann2022laion}\\ (For Quit-1M)\vspace{0.3em} \\  \textbf{Model/Framework:} \\ Whisper~\cite{radford2023robust}, GPT-3.5,  
\\ inaSpeechSegmenter~\cite{doukhan2018open},\\
langdetect~\cite{langdetect}} &
  \makecell[c]{Quilt-Net\cite{ikezogwo2024quilt}} &
  \makecell[c]{\textbf{Quilt-1M}\\ with 1 million \\
  image-text pair \\ with additional \\ data from LAION,\\ Twitter, and PubMed\\(Extension)
  } & \makecell[c]{\href{https://github.com/wisdomikezogwo/quilt1m}{\ding{51}}}
   \\
   \midrule

   OpenPath~\cite{huang2023visual} &
  \makecell[c]{Image-text \\ pairs} &
  \makecell[c]{ 208,414 \\ image–text pairs} &
  \makecell[c]{116,504 image–text\\ pairs 
from \\ Twitter posts,\\ 59,869 image–text \\ pairs from replies, \\32,041 image–text pairs\\ from LAION-5B}&
  \makecell[c]{Pathology 
image \\ classifier to exclude\\ non-pathology images,\\ CLIP image embeddings \\ with cosine similarity\\ to create PathLAION,\\ other hand-crafted \\ heuristics}& 
  \makecell[l]{\textbf{Dataset:}\\ LAION-5B~\cite{schuhmann2022laion}\vspace{0.3em} \\  \textbf{Model/Framework:} \\ 
   CLIP~\cite{radford2021learning}, langdetect~\cite{langdetect} } &
  \makecell[c]{PLIP~\cite{huang2023visual}} &
  \makecell[c]{\textbf{PathLAION} \\ 32,041 pathology \\ images from\\  the 
LAION-5B dataset\\ (Subset)
  } & \makecell[c]{\href{https://github.com/PathologyFoundation/plip}{\ding{51}}}
   \\
      \midrule

      \makecell[l]{Dataset from \\MI-Zero ~\cite{lu2023visual}} &
  \makecell[c]{Image-caption \\ pairs} &
  \makecell[c]{33,480 \\ image-caption pairs} &
  \makecell[c]{Publicly available \\ educational resources \\ combined with ARCH}&
  \makecell[c]{Hand-crafted algorithms \\ to filter out \\ non-pathology images
}& 
  \makecell[l]{\textbf{Dataset:} \\ 
  ARCH~\cite{gamper2021multiple}} &
  \makecell[c]{MI-Zero~\cite{lu2023visual}} &
  \makecell[c]{\textbf{---}} & \makecell[c]{\href{https://github.com/mahmoodlab/MI-Zero}{\ding{51}}}
   \\
      \midrule

   ARCH~\cite{gamper2021multiple} &
  \makecell[c]{Image-caption \\ pairs} &
  \makecell[c]{11,816 \\ image-caption \\ pairs} &
  \makecell[c]{PubMed  and pathology\\ textbooks  }&
  \makecell[c]{Hand-crafted \\ algorithms with \\tools like \\  PubMed Parser, \\ Pdffigures 2.0
}& 
  \makecell[l]{\textbf{Model/Framework:} \\ 
  PubMed Parser~\cite{Achakulvisut2020} \\ Pdffigures 2.0~\cite{clark2016pdffigures}} &
  \makecell[c]{\textbf{---}} &
  \makecell[c]{\textbf{---}} & \makecell[c]{\href{https://warwick.ac.uk/fac/cross_fac/tia/data/arch}{\ding{51}}}
   \\

    \midrule
   PathVQA~\cite{he2020pathvqa} &
  \makecell[c]{Image and \\ Q/A pairs} &
  \makecell[c]{4,998 images\\ 32,799 Q/A pairs} &
  \makecell[c]{Pathology textbooks,  \\Pathology Education \\ Informational Resource \\ (PEIR) digital library}&
  \makecell[c]{Hand-crafted \\ algorithms with \\tools like \\ PyPDF2, PDFMiner, \\Stanford CoreNLP \\ toolkit
}& 
  \makecell[l]{\textbf{Model/Framework:} \\ 
  PyPDF2, PDFMiner \\ Stanford CoreNLP toolkit~\cite{klein2003accurate}} &
  \makecell[c]{\textbf{---}} &
  \makecell[c]{16,465 open-ended\\ Q/A subset,\\ 16,334 close-ended \\ Q/A subset } & \makecell[c]{\href{https://huggingface.co/datasets/flaviagiammarino/path-vqa}{\ding{51}}}
   \\

   \bottomrule  
   
\multicolumn{9}{l}{\makecell[l]{\textbf{Availability}: GitHub repositories (except for ARCH~\cite{gamper2021multiple} and PathVQA~\cite{he2020pathvqa}; direct dataset source is linked for these two papers) with the associated paper are linked if they are open-sourced and accessible. \\ Some repositories only provide data generation and processing code (if the data is proprietary or requires an API) and some provide direct data sources through platforms such as Hugging Face Hub, Zenodo.\\\textbf{Other Utilized
Dataset/Models:} Dataset/Models/Framework used only in the data generation/processing part is mentioned. The downstream task datasets are not mentioned for this reason.}}  \\
   
\end{tabular}}
\label{tab:multi-modal_datasets}
\end{table*}

\subsection{Type of Datasets:}
The first component is the type of data which can be broadly classified into $5$ categories. Prov-Path~\cite{xu2024whole} fall into the last category which is dissimilar to the rest of the datasets. The dataset for Prov-Path utilizes WSI and the corresponding reports along with histopathology findings, 
cancer staging, genomics mutation profiles, etc. collected by Providence
Health System (PHS).

 The rest of the datasets can be sectioned into the remaining four categories. The image-caption/image-text pair category (dataset from
CONCH~\cite{lu2024visual}, PathGen-1.5M~\cite{sun2024pathgen}, Quilt-1M~\cite{ikezogwo2024quilt}, PathCap~\cite{sun2024pathasst}, OpenPath~\cite{huang2023visual}, dataset from
MI-Zero~\cite{lu2023visual}, ARCH~\cite{gamper2021multiple}) involves a low or medium-quality image and an associated piece of text for that image. This text can be a short caption with a description of the image or a more elaborate description. ARCH is the earliest dataset in this category that utilized PubMed and pathology textbooks to extract the texts. PathGen-1.5M is the latest dataset in this category, but unlike other datasets in this category, the images are patches extracted from WSIs. 
   \begin{figure}[htbp]
    \centering
\includegraphics[width=0.75\linewidth]{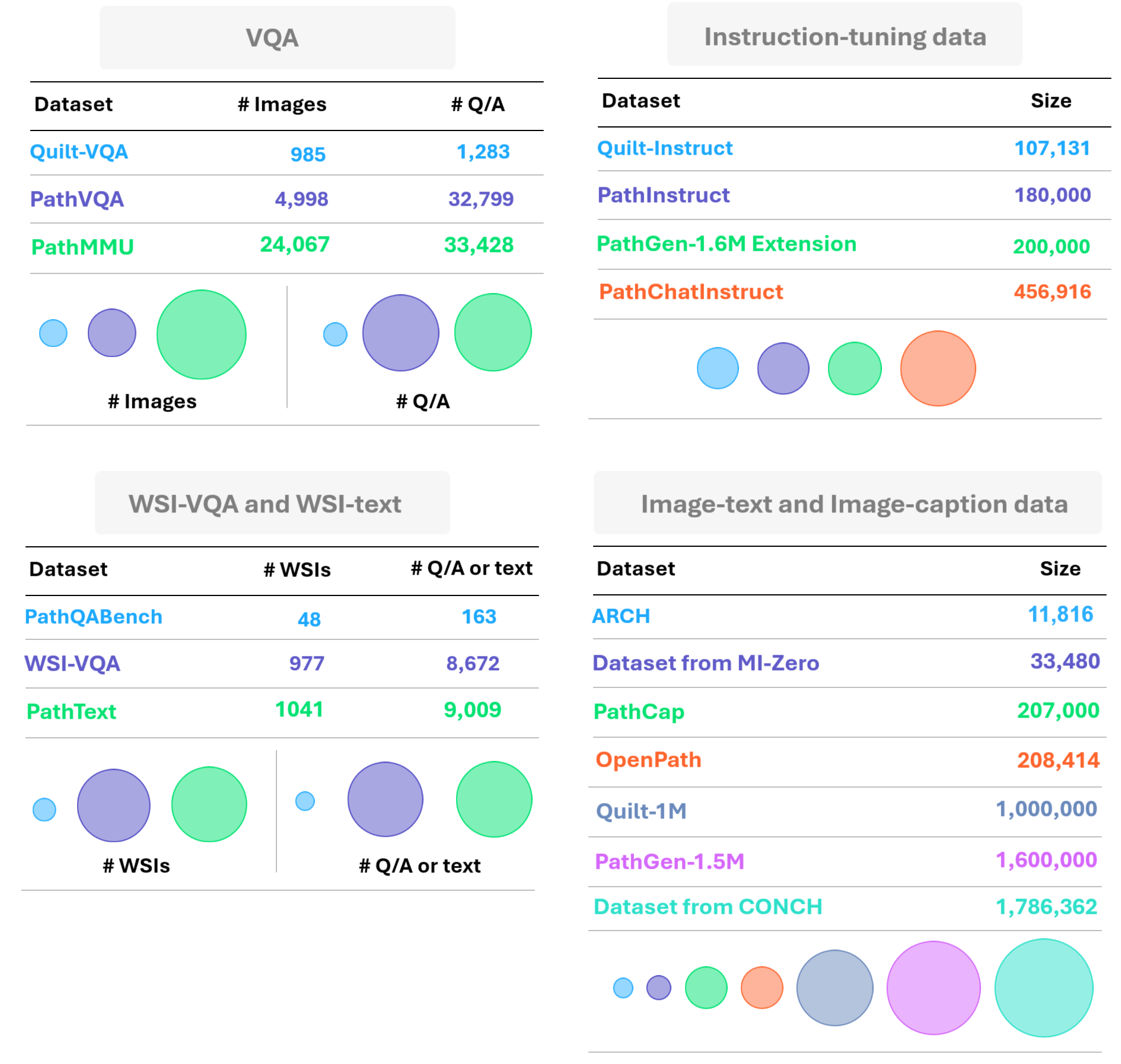}
    \caption{Comparison between the size of different multi-modal datasets. The size of the bubbles indicates the size of the data set (For visual clarity, the scale used for bubble size is the same within a specific group, but differs between groups).}
    \label{fig:multi-modal_size}
\end{figure}
 
 This is in contrast to other datasets in this category like OpenPath~\cite{huang2023visual} which is constructed with Twitter posts and replies and Quilt-1M~\cite{ikezogwo2024quilt} which is constructed with frames extracted from educational pathology videos. However, the curation and annotation process of this category is much easier as it can be automated with hand-crafted algorithms and heuristics. However, it comes with an inevitable noisy data as due to the automated process a lot of artifacts can be present in the data.

 The WSI VQA/text category (WSI-VQA~\cite{chen2024wsi}, PathQABench~\cite{lu2024multimodal}, PathText~\cite{chen2023mi}) contains question and answer pairs or texts associated with WSIs. The most common data source is the cancer genome atlas (TCGA)~\cite{weinstein2013cancer} which contains a large repository of WSI and patient report pairs. The VQA part can be of two types, close-ended question-answer pair and open-ended question-answer pair. Close-ended question-answer pairs are of a multiple-choice type or short-answer type which have pre-defined answers. On the other hand, open-ended question-answer pairs contain answers that are in natural language form. Among the datasets in this category, PathQABench is unique as it contains region of interest (ROI) annotation of WSIs performed by expert pathologists. It has two subsets as PathQABench-Public and PathQABench-Private. The former is publicly available as it was constructed with TCGA WSIs and reports, and the latter was constructed with in-house data. 

\begin{figure}[!ht]
    \centering
    \includegraphics[width=0.8\linewidth]{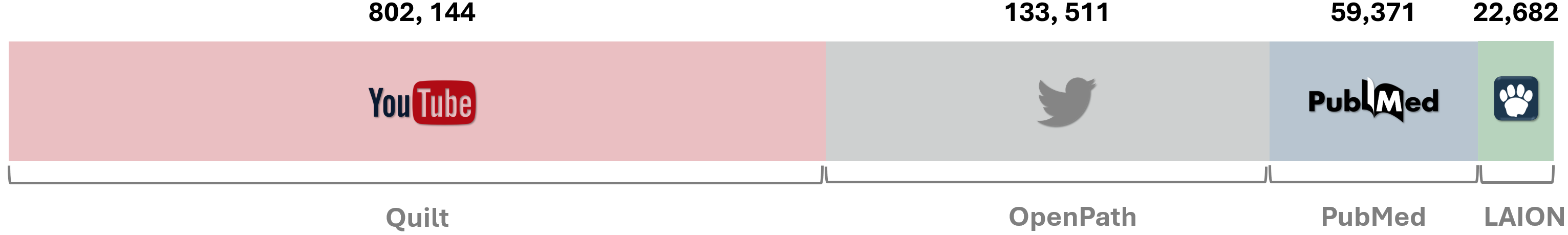}
    \caption{Subsets of the \textbf{Quilt-1M} dataset and the corresponding number of image-text pair for each subset.}
    \label{fig:subsets}
\end{figure}

 \begin{figure}[!ht]
     \centering
     \includegraphics[width=0.8\linewidth]{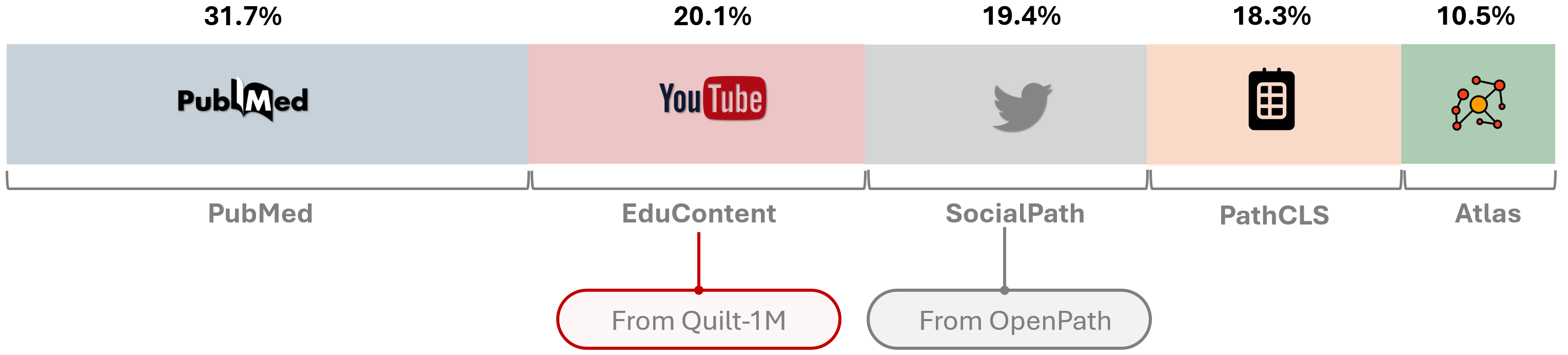}
     \caption{Subsets of the \textbf{PathMMU} dataset and proportion for each subset. The EduContent and SocialPath subset is sourced from Quilt-1M and OpenPath dataset.}
     \label{fig:subsetpathmmu}
 \end{figure}
 The next category, which is VQA (PathMMU~\cite{sun2024pathmmu}, Quilt-VQA~\cite{seyfioglu2024quilt}, PathVQA~\cite{he2020pathvqa}) is similar to the previous category as it also contains close-ended and open-ended question-answer pairs, but the associated images are not WSIs but rather low and medium-quality images. Among these datasets, PathVQA is the first research to curate a pathology-specific VQA dataset. PathMMU is the latest and largest dataset in this category and it also provides explainability annotations with each answer.
 Another category, which is the instruction-tuning dataset (Quilt-Instruct~\cite{seyfioglu2024quilt}, PathInstruct~\cite{sun2024pathgen}, PathChatInstruct~\cite{lu2024multimodal}, extension of PathGen-1.6M~\cite{sun2024pathgen}) is a unique kind of dataset, as this type of dataset is used to provide conversational ability to an existing multimodal model. The common workflow is that the instruction-tuning dataset is applied in the last phase to fine-tune an already trained VLFM or VLM. All of these datasets were created following the strategy mentioned in LLaVA~\cite{liu2024visual} or LLaVA-1.5~\cite{liu2024improved}. More details about the instruction-tuning phase are provided in section~\ref{sec:ins_tune} and section~\ref{sec:v_tune}.

A comparison of dataset sizes is shown in Fig.~\ref{fig:multi-modal_size}.

\subsection{Source of Data:}

 The second component to consider is the source of data which largely dictates the third component, the annotation and pre-processing. PubMed is a common data source containing pathology images and captions/text. However, the quality of the data is not as high as that of the TCGA repository that contains WSIs and corresponding pathological reports. Other high-quality data sources that contain WSIs and pathology reports are in-house proprietary datasets (PathQABench-Private, Prov-Path). A unique data source, OpenPath, contains pathology images and texts from Twitter posts and replies associated with a large pathology community.

 This data set was supplemented by pathology-specific data from the large-scale artificial intelligence open network (LAION) data repository~\cite{schuhmann2022laion}. Pathology textbooks and Atlas are also large knowledge sources that can be used to extract image caption/text pairs.  In a couple of recent research~\cite{ikezogwo2024quilt,seyfioglu2024quilt}, educational histopathology videos on YouTube are being used as the source of pathology image and text pair. However, curation of this kind of dataset requires a series of hand-crafted algorithms and many external tools. 
\begin{figure*}[!t]
    \centering
    \includegraphics[width=0.65\linewidth]{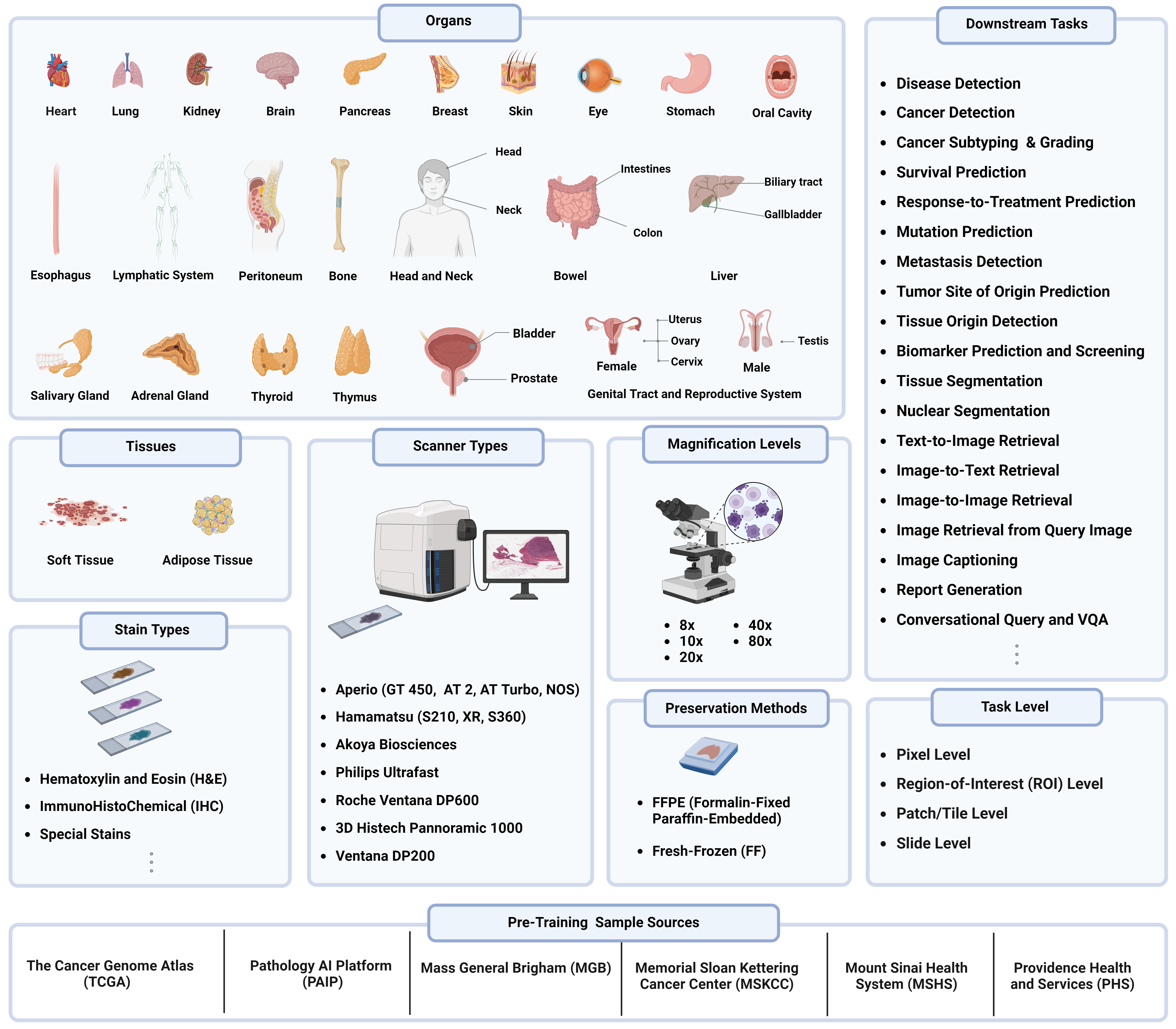}
    \caption{A high-level visualization of different factors of variability in terms of organs, stain types, scanner types, magnification levels, preservation methods, downstream tasks, task levels, pre-training tissue sample sources, etc in foundation models. The research that puts the most emphasis on ensuring variability are RudolfV~\cite{dippel2024rudolfv} and PLUTO~\cite{juyal2024pluto}. Among the tissue sample sources for pre-training,  TCGA and PAIP are publicly accessible and the rest are proprietary.}
    \label{fig:fm_diversity}
\end{figure*}
 Based on the above discussion, it is apparent that there is a trade-off between the quality and volume of the data. 

 Another key point is that most of these datasets contain other datasets as one of the subsets. An example of that is Quilt-1M which contains OpenPath, PubMed and LAION as subsets in addition to the proposed Quilt dataset. A visualization of Quilt-1M and its subsets is provided in Fig.~\ref{fig:subsets}. Another such example is the PathMMU dataset (shown in Fig.~\ref{fig:subsetpathmmu}) which contains $5$ different subsets, PubMed, SocialPath, EduContent, Atlas and PathCLS each containing data from different sources.

\subsection{Annotation and Pre-Processing:}

The third component is the annotation and pre-processing steps of data curation and generation. Among the surveyed articles, all employ a series of steps depending on the type of dataset and the pre-processing pipeline is unique for every article. However, some specific steps are similar if the data source is the same. For example, all articles that utilize PubMed as a data source use some kind of parsing process~\cite{Achakulvisut2020} to parse and extract figures and texts.  In addition, they employ light-weight classifiers and object detection architectures (YOLOv5~\cite{redmon2016you}, YOLOv6~\cite{li2022yolov6}, YOLOv7~\cite{wang2023yolov7}) to distinguish between pathology and non-pathology images, detect and separate subfigures, etc. Another common approach is to prompt LLMs to format and refine captions/text or structure extracted information according to a pre-defined template. These LLMs include generalized LLMs like GPT-4, GPT-3.5, ChatGPT and also specialized LLMs like BioGPT. Another widely used strategy is using a trained CLIP-based model and using cosine similarity as a metric to classify pathology and non-pathology images. 

Apart from the approaches mentioned above, there are a lot of other hand-crafted algorithms, heuristics and tools which are summarized in Table~\ref{tab:multi-modal_datasets}. 

\section{Foundation Model}\label{sec:foundation_models}

In this section, an overview of existing FMs in CPath is provided. First, the characteristics of FMs (section~\ref{sec:char_FM}) are provided to remove any ambiguity for the later sections. Next, pre-training workflow and typical pre-training schemes (section~\ref{pre-training}) are mentioned along with the instruction-tuning phase (section~\ref{sec:ins_tune}) which is becoming a widely adopted technique to provide FMs with conversational ability. Next, the downstream tasks and the associated downstream datasets (section~\ref{downstream}) for the FMs are outlined.

\subsection{Characteristics of FMs}\label{sec:char_FM}

A model can be classified as FM if it holds the following characteristics :

\begin{enumerate}
    \item The first characteristic that is common to all FMs is the SSPT. The data used in the pre-training phase do not have any explicit label or annotation. 
    \item The training goal of FMs is not to solve any specific task but rather to learn a general and rich representation space. For VFMs, it is a vision representation space and for VLFMs it is a vision-language representation space. The training of FMs is termed as ``pre-training" as in later stages further training is required to optimize for a specific task. 

    \item In CPath, FMs are trained using large and diverse datasets that encompass tissue samples from different organs and anatomic sites. In addition, some research~\cite{dippel2024rudolfv,juyal2024pluto} put effort into capturing diversity in terms of scanner types, magnification levels, stain types, preservation methods, etc.  The idea is to capture the representation of a wide range of tissue and disease types. In Fig.~\ref{fig:fm_diversity} a high-level visualization is provided highlighting different aspects of FMs.

    \item Another characteristic is the size of models that typically have parameters on the scale of millions. A huge amount of computing resources is put into training these models involving multiple GPUs. 
    
\end{enumerate}

As shown in Fig.~\ref{fig:timeline}, the FMs are sectioned into three separate categories. One category encompasses VFMs~\cite{zimmermann2024virchow,hua2024pathoduet,vorontsov2024foundation,hoptimus0,dippel2024rudolfv,nechaev2024hibou,yang2024foundation,juyal2024pluto,zhou2024knowledge,chen2024towards,campanella2024computational,filiot2023scaling,kang2023benchmarking,wang2022transformer} in CPath, the second category encompasses VLFMs~\cite{ma2024towards,lu2024multimodal,xu2024whole,shaikovski2024prism,sun2024pathasst,lu2024visual,huang2023visual} and the last category utilizes these FMs by providing a benchmark, framework
or adapting existing FMs~\cite{lu2024multiple,mallya2024benchmarking,lu2024pathotune,campanella2024clinical,zheng2024benchmarking,yin2024prompting,aswolinskiy2024impact,aben2024towards,ferber2024context,alfasly2024foundation,roth2024low,lai2023domain}.
\begin{figure*}[!t]
    \centering
    \includegraphics[width=0.73\linewidth]{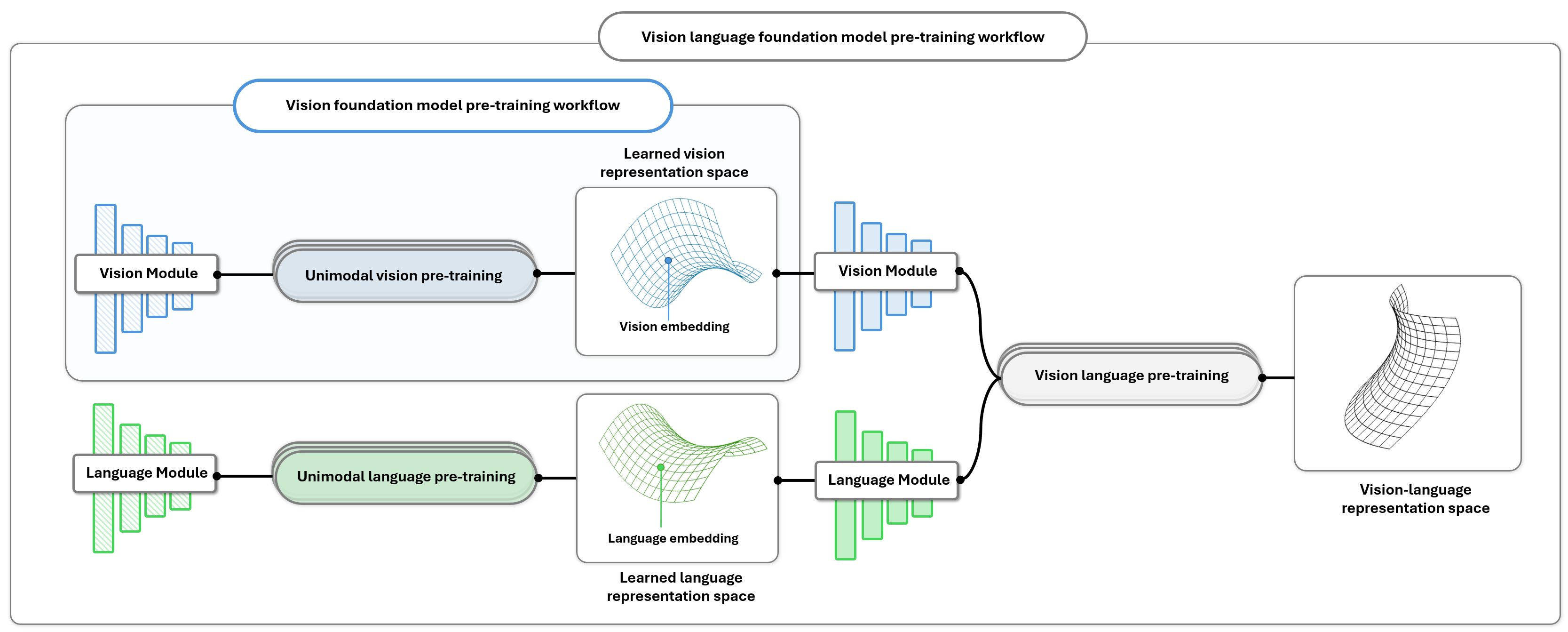}
    \caption{Typical pre-training workflow in FMs. In general it starts with a vision module and a language module with randomly initialized weights or pre-set weights. VFMs only go through unimodal vision pre-training to learn a vision representation space. On the other hand, VLFMs can optionally go through unimodal pre-training for their vision and language modules. }
    \label{fig:vlm_pretraining}
\end{figure*}
\subsection{Pre-training Workflow and Strategies}\label{pre-training}

The typical pre-training workflow of FMs is shown in Fig.~\ref{fig:vlm_pretraining} which provides a high-level visualization of different phases in the workflow. 

In this single diagram, both pre-training strategy for VFM and VLFM is shown. For VFMs, a vision module with some initial weight is leveraged in pre-training. The term ``vision module" is used to generally represent a wide range of vision architectures and also modified versions of these architectures with additional layers. The most common architecture is variants of vision image transformers (ViTs)~\cite{dosovitskiy2020image} which includes ViT-S, ViT-B, ViT-L, ViT-G and ViT-H. Though there are specialized architectures proposed in BEPH~\cite{yang2024foundation} and Prov-GigaPath~\cite{xu2024whole} which uses BEiTv2~\cite{peng2022beit} and GigaPath architecture, respectively. Note that, some models initialize the architectures with ImageNet~\cite{imagenet15russakovsky} weights to get an initial vision representation space which can be transformed through pre-training. In Table~\ref{tab:pretrain_strategy}, a summary of pre-training strategies in different phases is outlined. 

\begin{table}[!ht]
\caption{Different pre-training phases and Corresponding SSPT Strategies}
\centering
\renewcommand{\arraystretch}{1.5}
\setlength{\tabcolsep}{10pt}
\resizebox{0.85\linewidth}{!}{\begin{tabular}{@{}lcc@{}}
\toprule
\textbf{Pre-training phase}                   & \multicolumn{2}{c}{\textbf{Strategy}}      \\ \midrule
\multirow{3}{*}{\makecell[l]{Unimodal vision\\ pre-training}} & \makecell[c]{Self-Distillation}          & DINO~\cite{caron2021emerging}, DINOv2~\cite{oquab2023dinov2} \\ \cmidrule(l){2-3} 
                                              & Contrastive Learning        & MoCov2~\cite{chen2020improved}, MoCov3~\cite{chen2021empirical}      \\ \cmidrule(l){2-3} 
                                              & \makecell[c]{Masked Image \\ Modeling (MIM)} & \makecell[c]{MAE~\cite{he2022masked},\\ iBOT (MIM+Self distillation)~\cite{zhou2021ibot}}          \\ \midrule
\makecell[l]{Unimodal language \\ pre-training}                & \multicolumn{2}{c}{No Specific Strategy}                   \\ \midrule
\makecell[l]{Vision-language \\ pre-training}                  & \multicolumn{2}{c}{ CLIP~\cite{radford2021learning}, CoCa~\cite{yu2022coca}}             \\ \bottomrule
\end{tabular}}
\label{tab:pretrain_strategy}
\end{table}
\begin{table*}[!ht]
\caption{Summary of Vision Pre-Training Strategy and Vision Pre-Training Dataset of Foundation Models in Pathology}
\centering
\renewcommand{\arraystretch}{2.5}
\setlength{\tabcolsep}{5pt}
\resizebox{\linewidth}{!}{
\begin{tabular}{llcccllllllcc}
\toprule
\multirow{2}{*}{\textbf{Model}} &
   &
  \multicolumn{3}{c}{\textbf{Vision Pre-Training Strategy}} &
   &
  \multicolumn{4}{c}{\textbf{Vision Pre-Training Dataset}} &
   &
  \multirow{2}{*}{\textbf{\makecell[c]{Type of \\ Foundation \\ Model }}} &
  \multirow{2}{*}{\makecell[c]{\textbf{Availability}\\(\textit{Linked})}} \\ \cline{3-5} \cline{7-10}
 &
   &
  \textbf{Category} &
  \textbf{Approach} &
  \textbf{Architecture} &
   &
  \textbf{Source} &
  \textbf{Size} &
  \textbf{Stain} &
  \textbf{\makecell[l]{Additional \\ Information}} &
   &
   &
   \\ \cline{1-1} \cline{3-5} \cline{7-10} \cline{12-13}  

   GPFM~\cite{ma2024towards}&
   & \makecell[c]{\makecell[c]{MIM + Self-Distillation+ \\ (Expert Knowledge Distillation)}}&
  \makecell[c]{Custom}&
\makecell[c]{Custom architecture\\ that uses \\ other FMs \\ including \\ UNI, Phikon, CONCH} &
   &
  \makecell[l]{$33$ public datasets \\ including \\ TCGA, GTExPortal\\ PAIP, CPTAC, etc} &
  \makecell[l]{\textbf{Slides:} 86,104\\ \textbf{Tiles:}  190,000,000} &
 \makecell[l]{\textbf{---}
} &
   \makecell[l]{47 data sources
\\ 34 major tissue types}  &
   &
  \makecell[c]{Combines \\ Knowledge of \\ VFMs and VLFMs} &  
  \href{https://github.com/birkhoffkiki/GPFM}{\ding{51}} \\ 

  \midrule
  \midrule
  
\makecell[l]{Virchow 2~\cite{zimmermann2024virchow}\\ \\ Virchow 2G}&
   &
   \makecell[c]{Self-Distillation} &
  DINOv2 &
  \makecell[c]{ViT-H \\  \\ ViT-G}  &
   &
  \makecell[l]{Memorial Sloan\\ Kettering \\ Cancer Center \\(MSKCC)\\ + Institutions worldwide} &
  \makecell[l]{\textbf{Slides:} 3,134,922 \\ \textbf{Tiles:} ---} &
  \makecell[l]{H\&E\\IHC}
   &
  \makecell[l]{225,401 patients\\ 493,332 cases \\ 871,025 specimens}  &
   &
  Vision &
  \href{https://huggingface.co/paige-ai/Virchow2}{\ding{51}}\\
  \midrule

 PathoDuet~\cite{hua2024pathoduet}&
   & \makecell[c]{Contrastive \\ Learning}&
  MoCov3&
  \makecell[c]{ViT-B} &
   &
 \makecell[l]{TCGA} &
  \makecell[l]{\textbf{Slides:} 11,000\\ \textbf{Tiles:}  13,166,437} &
 \makecell[l]{H\&E 
} &
  \makecell[l]{\textbf{---}}   &
   &
  Vision &  
  \href{https://github.com/openmedlab/PathoDuet}{\ding{51}} \\ 
  \midrule  
  
Virchow~\cite{vorontsov2024foundation}&
   &
   \makecell[c]{Self-Distillation} &
  DINOv2 &
  ViT-H &
   &
  \makecell[l]{Memorial Sloan \\ Kettering \\ Cancer Center \\(MSKCC)} &
  \makecell[l]{\textbf{Slides:} 1,488,550 \\ \textbf{Tiles:} 2 billion} &
  H\&E
   &
  \makecell[l]{17 organs\\ 119,629 patients\\ 208,815 cases \\ 392,268 specimens}  &
   &
  Vision &
  \href{https://huggingface.co/paige-ai/Virchow}{\ding{51}}\\
  \midrule
RudolfV~\cite{dippel2024rudolfv}&
   &
  \makecell[c]{Self-Distillation} &
  DINOv2 &
  ViT-L &
   &
 \makecell[l]{108,433 slides \\ from 15 labs \\across the\\ EU and US \\ +  26,565 slides \\ from TCGA} &
  \makecell[l]{\textbf{Slides:} 133,998 \\ \textbf{Tiles:} 1.2 billion } &
  \makecell[l]{H\&E (68\%)\\ + IHC (15\%)\\+ other (17\%)}
  &
 \makecell[l]{14 organs \\ 15 lab \\ 58 tissue types\\ 129 stains \\ 6 scanner types\\ FFPE and FF tissue
}  &
   &
  Vision &
  \textbf{---}\\
  \midrule

Hibou~\cite{nechaev2024hibou}&
   &
  \makecell[c]{Self-Distillation} &
  DINOv2 &
  \makecell[c]{ViT-B \\ (for Hibou-B)\\   ViT-L \\ (for Hibou-L) } &
   &
 \makecell[l]{Proprietary \\ Data} &
  \makecell[l]{\textbf{Slides:} 936,441 H\&E \\ +202,464 non-H\&E \\+ 2,676 cytology slides \\  \textbf{Tiles:}   1.2 billion 
patches \\ for Hibou-L, 512 million \\ patches  for Hibou-B } &\makecell[l]{H\&E \\ + non H\&E} 
   &
\makecell[l]{306,400 cases
}  &
   &
  Vision &  
  \makecell[c]{\href{https://github.com/HistAI/hibou}{\ding{51} }}\\
  \midrule

 BEPH~\cite{yang2024foundation}&
   &
 MIM &
  \textbf{---} &
  \makecell[c]{BEiTv2~\cite{peng2022beit} \\ with VQ-KD \\ autoencoder \\ and ViT-B encoder} &
   &
 \makecell[l]{TCGA} &
  \makecell[l]{\textbf{Slides:} 11,760 \\ \textbf{Tiles:}  11,774,353 } &
  \makecell[l]{H\&E } 
   &
    \makecell[l]{32 cancer types }&
   &
  Vision &  
  \href{https://github.com/Zhcyoung/BEPH}{\ding{51}} \\
  \midrule

PLUTO~\cite{juyal2024pluto}&
   &
  \makecell[c]{Self-Distillation+MIM} &
  \makecell[c]{DINOv2 variation \\ + MAE}  &
  \makecell[c]{ViT-S variant \\ FlexiViT-S~\cite{beyer2023flexivit}} &
 &
 \makecell[l]{TCGA\\+ Proprietary data \\from PathAI} &
  \makecell[l]{\textbf{Slides:}  195 million \\ \textbf{Tiles:} 158,000 } &
  \makecell[l]{H\&E\\+ IHC\\+ Other stains}
  &
 \makecell[l]{ More than 30 diseases\\ 
               More than 12 scanners \\ 
               More than 100 stain types\\ 
               More than 4M pathologist\\ pixel-level  annotation
}  &
   &
  Vision &
  \textbf{---}\\
  \midrule

 UNI~\cite{chen2024towards}&
   & \makecell[c]{Self-Distillation} &
  \makecell[c]{DINOv2\\ (also MoCov3 \\ for comparison)}&
  \makecell[c]{ViT-L} &
   &
 \makecell[l]{Proprietary Data from \\ Massachusetts\\  
General  Hospital (MGH), \\ Brigham and Women’s \\ Hospital (BWH), \\ Genotype–Tissue \\ Expression (GTEx)\\ consortium} &
  \makecell[l]{\textbf{Slides:} 100,426\\ \textbf{Tiles:}  over 100 million} &
 \makecell[l]{H\&E 
} &
  \makecell[l]{20 major tissue \\types}   &
   &
  Vision &  
  \href{https://github.com/mahmoodlab/UNI}{\ding{51}} \\
  \midrule 

 3B-CPath~\cite{campanella2024computational}&
   & \makecell[c]{MIM \\and \\ Self-Distillation}&
  MAE, DINO&
  \makecell[c]{ViT-S, \\ ViT-L} &
   &
 \makecell[l]{Mount Sinai 
Health \\ System (MSHS)} &
  \makecell[l]{\textbf{Slides:} 423,600\\ \textbf{Tiles:}  3.25 billion} &
 \makecell[l]{H\&E 
} &
  \makecell[l]{3 anatomic site\\ 2 institutions}   &
   &
  Vision &  
  \href{https://github.com/fuchs-lab-public/OPAL/tree/main/SSL_benchmarks}{\ding{51}} \\
  \midrule 

Phikon~\cite{filiot2023scaling}&
   & \makecell[c]{MIM+Self-Distillation}&
  iBOT&
  \makecell[c]{ViT-S, \\ ViT-B, \\ ViT-L} &
   &
 \makecell[l]{TCGA \\ (Three variants \\ TCGA-COAD,\\ PanCancer4M,\\
PanCancer40M)} &
  \makecell[l]{\textbf{Slides:} 6,093  \\ \textbf{Tiles:}  43,374,634 \\ (For PanCancer40M)} &
 \makecell[l]{H\&E
} &
  \makecell[l]{16 cancer types \\ 13 anatomic sites \\  5,558
patients \\ (For PanCancer40M)}   &
   &
  Vision &  
  \href{https://github.com/owkin/HistoSSLscaling}{\ding{51}} \\ 

\midrule
CTransPath~\cite{wang2022transformer}&
   & \makecell[c]{Contrastive \\ 
Learning }&
 \makecell[c]{MoCov3 variation \\(SRCL: semantically \\ relevant \\ contrastive
learning )}&
  \makecell[c]{Swin \\ Transformer} &
   &
 \makecell[l]{TCGA \\+ PAIP} &
  \makecell[l]{\textbf{Slides:} 32,220   \\ \textbf{Tiles:} 15.6 million } &
 \makecell[l]{H\&E
} &
  \makecell[l]{32 cancer types \\ 25 anatomic sites }   &
   &
  Vision &  
  \href{https://github.com/Xiyue-Wang/TransPath}{\ding{51}} \\ 
  
\midrule
\midrule

  Prov-GigaPath~\cite{xu2024whole}&
   & \makecell[c]{MIM \\ and \\ Self-Distillation}&
  \makecell[c]{MAE, DINOv2}&
  \makecell[c]{GigaPath~\cite{xu2024whole} \\ (constructed \\ with ViT and  \\LongNet~\cite{ding2023longnet})} &
   &
 \makecell[l]{Providence\\ Health System (PHS)} &
  \makecell[l]{\textbf{Slides:}  171,189\\ \textbf{Tiles:}  1.3 billion} &
 \makecell[l]{H\&E, \\ IHC 
} &
  \makecell[l]{30,000 patients\\28 cancer centers\\31 major tissue types}   &
   &
  \makecell[c]{Vision\\Language} &  
  \href{https://github.com/prov-gigapath/prov-gigapath}{\ding{51}} \\
  \midrule

  CONCH~\cite{lu2024visual}&
   & \makecell[c]{MIM + Self-Distillation}&
  \makecell[c]{iBOT}&
  \makecell[c]{ViT-B \\ backbone in \\ image encoder} &
   &
 \makecell[l]{In-house\\ dataset} &
  \makecell[l]{\textbf{Slides:}  21,442\\ \textbf{Tiles:}  1.2 million} &
 \makecell[l]{H\&E,\\ IHC, \\ Masson’s \\trichrome, \\Congo red
} &
  \makecell[l]{350 cancer subtypes}   &
   &
  \makecell[c]{Vision \\ Language} &  
  \href{https://github.com/mahmoodlab/CONCH}{\ding{51}} \\
\bottomrule
\end{tabular}}
\label{tab:vision_pre_training}
\end{table*}

The term ``unimodal" is used to signify the pre-training phase utilizing a single modality out of vision and language. This is inherently different from vision-language pre-training strategies which involve both vision and language modalities to learn a joint vision-language representation space. In the unimodal vision pre-training phase there are three strategies commonly used in CPath which are self-distillation, contrastive learning and masked image modeling (MIM) approach.  Each approach has its own advantages and disadvantages.

\begin{figure}[!ht]
    \centering
    \includegraphics[width=0.55\linewidth]{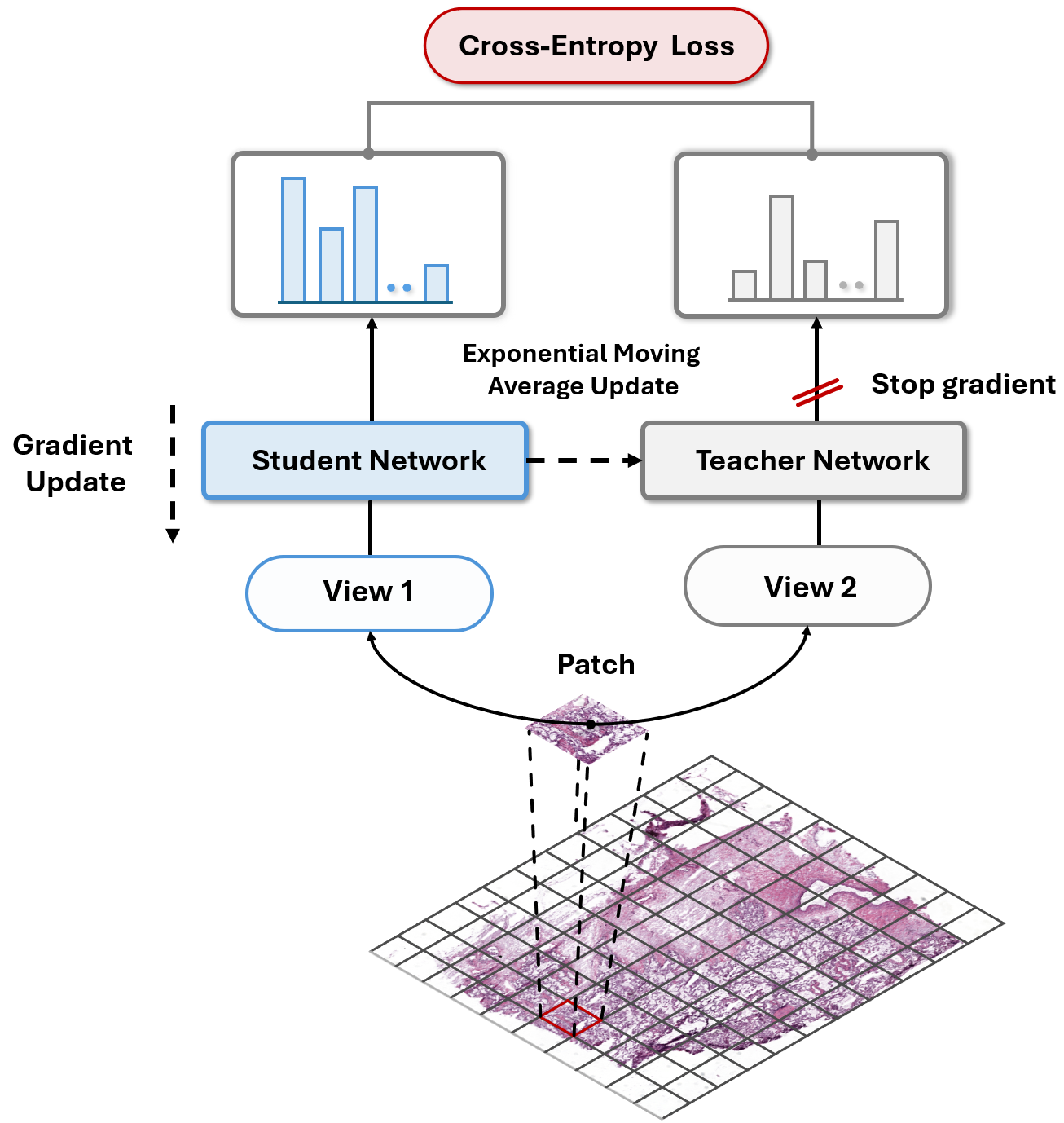}
    \caption{Visualization of the self-distillation approach for unimodal vision pre-training scheme.}
    \label{fig:dino}
\end{figure}
 The work done in Lunit\cite{kang2023benchmarking} benchmarks DINO~\cite{caron2021emerging}, MoCov2~\cite{chen2020improved}, SwaV~\cite{caron2020unsupervised}, Barlow Twins~\cite{zbontar2021barlow} but concludes there is no clear best SSPT scheme. However, the authors conclusively prove that SSPT always outperforms ImageNet initialization.
\subsubsection{Unimodal Vision Pre-training}
\begin{table*}
\centering
\caption{Summary of Vision-Language Pre-Training and Instruction Tuning Phase of Foundation Models in Pathology}
\centering
\renewcommand{\arraystretch}{2}
\setlength{\tabcolsep}{6pt}
\resizebox{0.9\linewidth}{!}{
\begin{tabular}{llccclcclllcc}
\hline
\multirow{2}{*}{\textbf{Model}} &
   &
  \multicolumn{3}{c}{\textbf{Vision Language Modules}} &
   &
  \multicolumn{4}{c}{\textbf{Vision-Language Pre-Training Phase}} &
   &
  \multicolumn{2}{c}{\textbf{Instruction Tuning Phase}} \\ \cline{3-5} \cline{7-10} \cline{12-13} 
 &
   &
  \textbf{Vision Module} &
  \textbf{Language Module} &
  \textbf{\makecell[c]{Additional \\Layer/Module}} &
   &
  \textbf{\makecell[c]{Architecture / \\ Framework}} &
  \multicolumn{3}{c}{\textbf{Pre-Training Process}} &
   &
  \textbf{\makecell[c]{Instruction\\ Tuning}} &
  \textbf{Process} \\ \cline{1-1} \cline{3-5} \cline{7-10} \cline{12-13} 

      \makecell[l]{PathChat~\cite{lu2024multimodal}}&
   &
  \makecell[c]{UNI~\cite{chen2024towards} as vision \\ encoder backbone} &
  \makecell[c]{Pre-trained \\ Llama 2~\cite{touvron2023llama}\\ LLM which is a \\ decoder-only \\ transformer-based \\ auto-regressive \\ model}  &
 \makecell[c]{A multimodal projector \\ module to connect \\ the outputs of the\\ vision module to\\ language module by\\ projecting the visual tokens\\ to the same dimension \\ as the LLM’s embedding \\space for 
text tokens} &
   &
  \makecell[c]{CoCa} &
  \multicolumn{3}{c}{\makecell[c]{Vision-language\\ pre-training according \\ to CoCa framework \\ with CONCH dataset}} &
   &
  \ding{51} &
  \makecell[c]{LLaVA-1.5~\cite{liu2024improved}\\ training approach\\ \textbf{First phase:} Only the \\ parameters of multi-modal \\ projector is updated \\ \textbf{Second phase:} Fine-tuned\\ with instruction-following\\ data}\\
  \midrule

\makecell[l]{PRISM~\cite{shaikovski2024prism}}&
   &
  \makecell[c]{Virchow as\\  tile encoder, \\Perceiver~\cite{jaegle2021perceiver} as \\ slide-encoder} &
  \makecell[c]{BioGPT \\ first 12 layers}  &
 \makecell[c]{BioGPT \\ last 12 layers as \\ vision-language\\
decoder} &
   &
 CoCa&
  \multicolumn{3}{c}{\makecell[c]{Trained using\\ contrastive loss and \\ generative/captioning loss}} &
   &
  \ding{55} &
  \makecell[c]{\textbf{---}}\\

  \midrule
  \makecell[l]{PathCLIP\\ and PathAsst~\cite{sun2024pathasst}}&
   &
  \makecell[c]{Image encoder with \\ ViT-B} &
  \makecell[c]{Text transformer \\ as the 
text encoder \\ with modifications \\ mentioned in~\cite{radford2019language} }  &
 \makecell[c]{Fully-connected layer \\ after vision encoder \\ to map the image \\ embedding space to the \\ corresponding \\ language embedding} &
   &
 CLIP &
  \multicolumn{3}{c}{\makecell[c]{Fine-tuning of \\ CLIP model through \\ contrastive learning using \\PathCap dataset}} &
   &
  \ding{51} &
  \makecell[c]{\textbf{First phase:} Detailed \\
description-based part is used \\ to train the fully-connected \\ layer connected to the vision\\ encoder. \\ \textbf{Second phase:} fine-tuned with\\ instruction-following  data \\ via next word prediction}\\
  \midrule
  
CONCH~\cite{lu2024visual}&
   &
  \makecell[c]{\\ An image encoder with\\ ViT-B  backbone \\ with 12 transformer \\ layers, 12 attention \\ heads followed \\ by two 
attentional pooler \\ modules} &
  \makecell[c]{A GPT-style \\text encoder\\with 12 
transformer \\ layers}  &
 \makecell[c]{A GPT-style \\multimodal decoder \\ with 12 
transformer \\ layers} &
   &
  CoCa &
  \multicolumn{3}{c}{\makecell[c]{Visual-language\\  pre-training with \\ image–text  contrastive \\ loss  and the \\ captioning loss according \\ to CoCa framework}} &
   &
  \ding{55} &
  \textbf{---}\\
  \midrule

  PLIP~\cite{huang2023visual}&
   &
  \makecell[c]{An image encoder with \\ ViT-B} &
  \makecell[c]{ A text transformer \\ as the 
text encoder \\ with modifications \\ mentioned in~\cite{radford2019language} }  &
 \makecell[c]{\textbf{---}} &
   &
 CLIP &
  \multicolumn{3}{c}{\makecell[c]{Fine-tuning of \\ CLIP model through \\ contrastive learning using \\OpenPath dataset}} &
   &
  \ding{55} &
  \textbf{---}\\

  \bottomrule
\end{tabular}}
\label{tab:vision_language_pretraing}
\end{table*}
\forestset{
  my tier/.style={
    tier/.wrap pgfmath arg={level##1}{level()},
  },
}
 The self-distillation with no label approach uses a student-teacher network to learn a rich vision representation space as shown in Fig.~\ref{fig:dino}. From a single image patch, two different views are generated by applying an augmentation sampled from a set of possible augmentations (color jittering, Gaussian blur, polarization, etc). The generated output of both the networks is utilized to compute a cross-entropy loss which is then used to update the parameters of the student network. The parameter of the teacher network is then updated through an exponential moving average (EMA) of the student network parameters. Among the surveyed articles Virchow2~\cite{zimmermann2024virchow}, Virchow~\cite{vorontsov2024foundation}, RudolfV~\cite{dippel2024rudolfv}, Hibou~\cite{nechaev2024hibou} and UNI~\cite{chen2024towards} use DINOv2~\cite{oquab2023dinov2} as the self-distillation approach. PLUTO~\cite{juyal2024pluto} takes a unique approach by integrating MAE and a Fourier loss term to get a custom variation of DINOv2. 

\begin{figure}[!ht]
    \centering
    \includegraphics[width=0.8\linewidth]{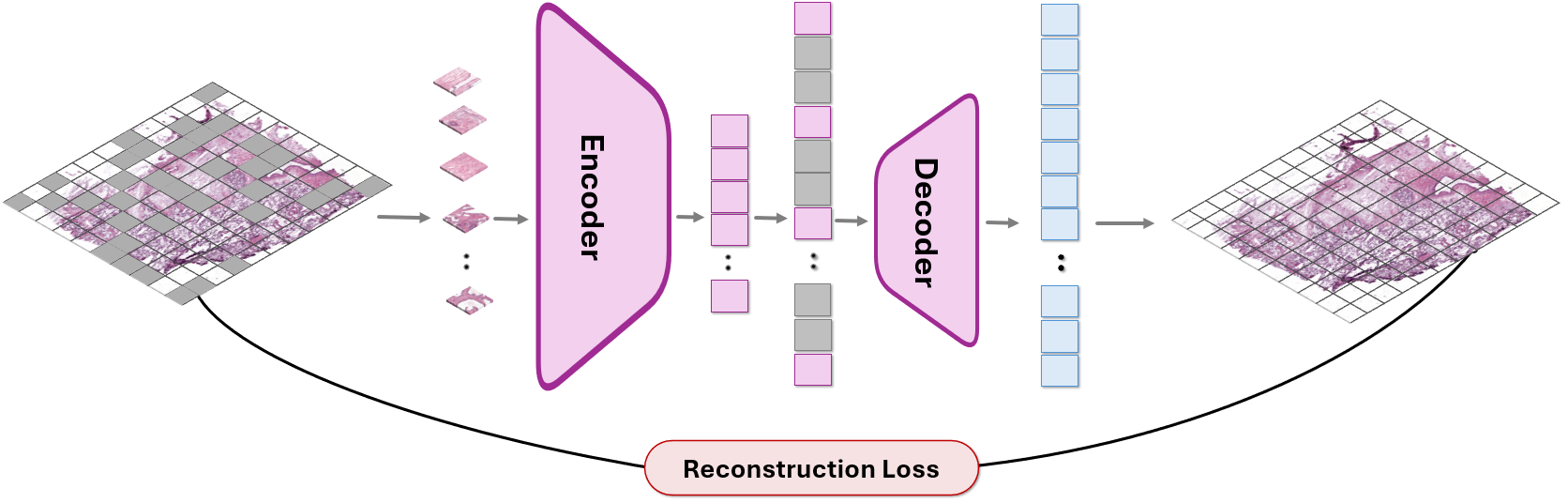}
    \caption{Visualization of the masked image modeling approach for unimodal vision pre-training scheme.}
    \label{fig:mim}
\end{figure}

The second popular SSPT approach is the MIM (visualized in Fig.~\ref{fig:mim}) which has variants like MAE~\cite{he2022masked} and iBOT~\cite{zhou2021ibot} used in FMs. In the MAE approach, randomly selected high portions of the image are masked out and the patches that are not masked out are passed through an encoder which generates latent representations of those patches. 

Then those representations are passed through a decoder along tokens of masked out regions to reconstruct the image and reconstruction loss is used to train the model.  The iBOT approach also uses MIM but adds self-distillation technique by leveraging a student-teacher network. The teacher network works as an online tokenizer and the student network learns to predict masked patches with the help of distilled knowledge of the teacher network.

Among the surveyed articles 3B-CPath~\cite{campanella2024computational} and Prov-GigaPath~\cite{xu2024whole} use MAE but in conjunction with DINO and DINOv2, respectively. Research works utilizing iBOT approach include Phikon~\cite{filiot2023scaling} and CONCH~\cite{lu2024visual}.

\begin{figure}[!ht]
    \centering
    \includegraphics[width=0.6\linewidth]{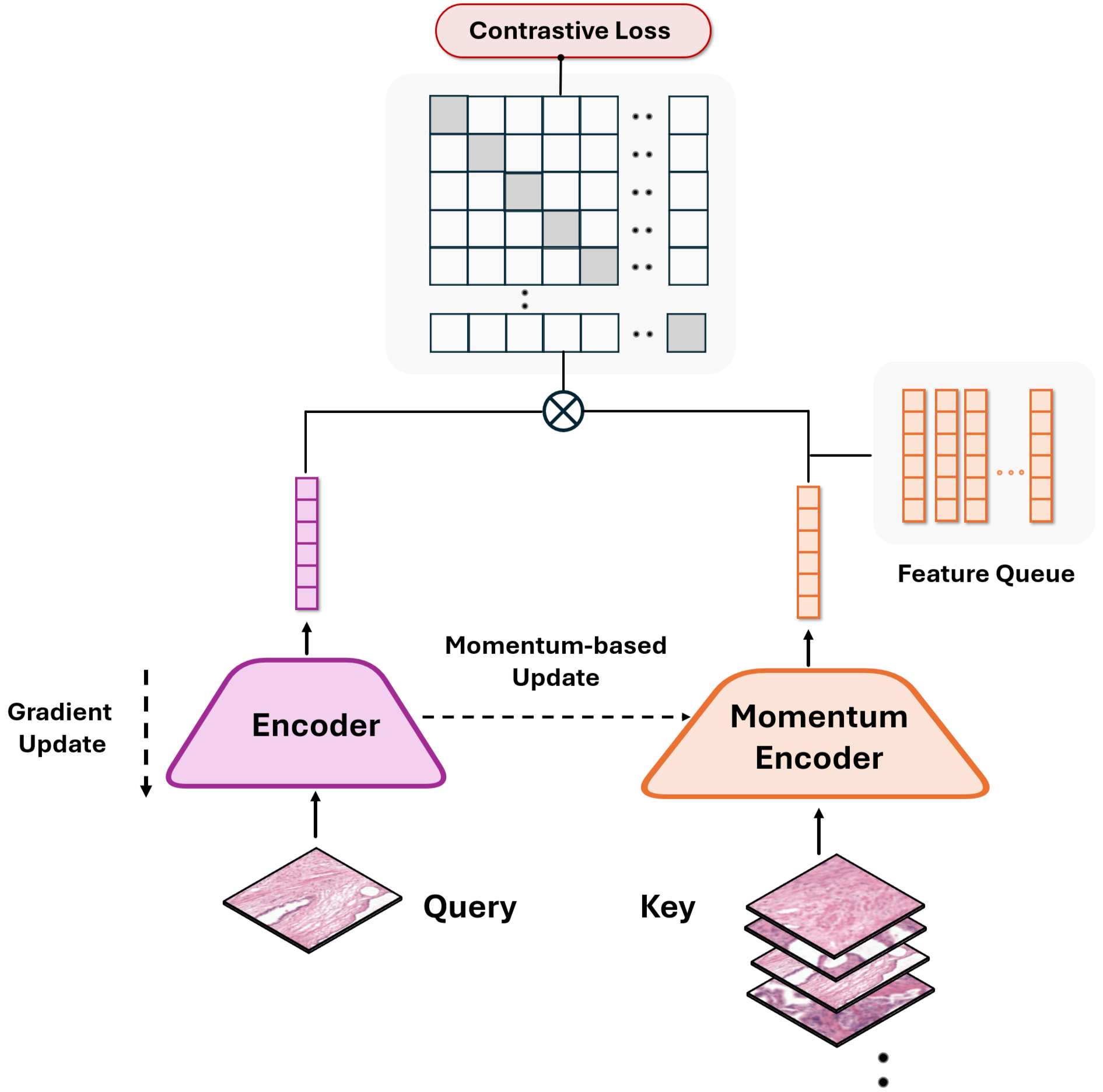}
    \caption{Visualization of the MoCo approach for unimodal vision pre-training scheme.}
    \label{fig:Moco}
\end{figure}

Another SSPT approach that is comparatively less popular in FMs for CPath is the contrastive learning framework proposed in MoCo~\cite{he2020momentum} (visualization in Fig.~\ref{fig:Moco}). Over the years variations of the proposed approach in MoCo in terms of architectural change and training blueprint have been done and MoCov2~\cite{chen2020improved} and MoCov3~\cite{chen2021empirical} are the results of that.

Like the self-distillation approach, MoCo also utilizes two models; one is an encoder (with query patch as input) and the second one is a special momentum encoder (with key patches as input). The embeddings generated through the encoder and the momentum encoder are used to compute a similarity score which in turn is used for contrastive loss computation. The computed loss is used in backpropagation to update the parameters of the encoder. The parameter of the momentum encoder is updated through a momentum-based update rule that utilizes the parameters of the encoder. 
Another innovation of MoCo was introducing a feature queue which ensures a large set of negative samples without holding the entire dataset in memory. This also ensures the negative samples do not go stale over the training period and the model sees a diverse set of negative samples to learn a better vision representation. Among the surveyed articles PathoDuet~\cite{hua2024pathoduet} uses MoCov3, UNI~\cite{chen2024towards} uses MoCov3 for comparison and CTransPath~\cite{wang2022transformer} performs a unique variation of MoCov3 called semantically relevant contrastive learning (SRCL).
GPFM~\cite{ma2024towards} is distinct from other approaches as it includes a novel expert knowledge distillation in addition to MIM and self-distillation by utilizing existing FMs UNI, Phikon and CONCH. 

\subsubsection{Unimodal Language Pre-Training}
Most VLFMs do not perform large-scale unimodal language pre-training. Instead, they rely on already pre-trained in-domain (trained with medical corpus) LLMs like BioGPT or other LLMs like Llama 2, GPT-variants, etc. One exception is CONCH which pre-trains the language module with $550,000$ 
surgical pathology reports from Massachusetts General Hospital and 
over $400,000$ select histopathology-relevant PubMed abstracts.

\subsubsection{Vision-Language Pre-training}

After unimodal pre-training for the image module and text module, VLFMs might go through vision language SSPT to align or learn a joint vision-language representation space.

\begin{figure}[!h]
    \centering
    \includegraphics[width=0.45\linewidth]{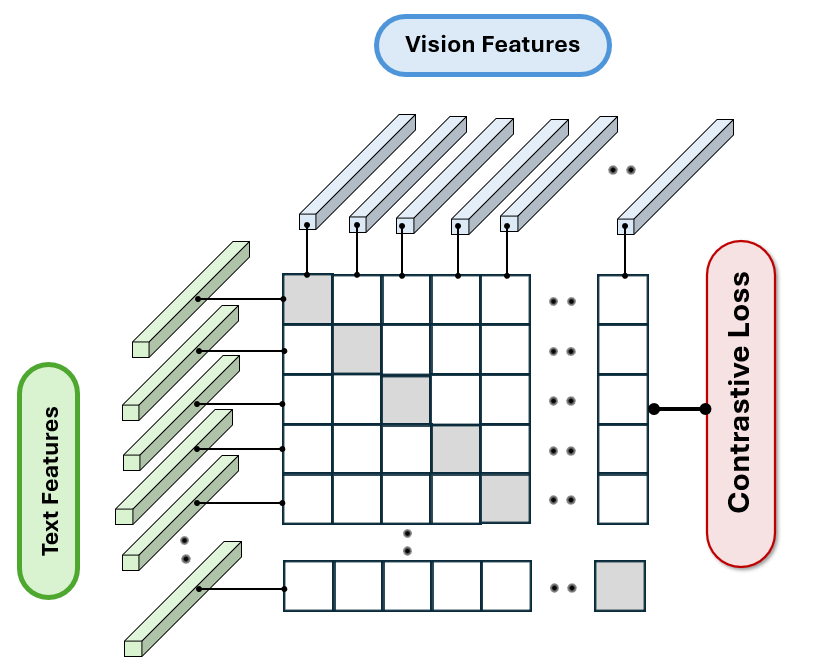}
    \caption{Visualization of the CLIP approach for the vision-language pre-training scheme.}
    \label{fig:clip}
\end{figure}

The most popular vision-language SSPT approach in CPath is the CLIP approach shown in Fig.~\ref{fig:clip}. 

The inputs to CLIP are a batch of image features encoded by an image encoder and a batch of text features encoded by a text encoder. The model is trained to pull together the corresponding image feature and text feature in the representation space and push apart the rest of the features i.e. utilizing a contrastive objective. In other words, maximizing the similarity (specifically cosine similarity) between the corresponding image and text pair while minimizing the cosine similarity with other non-matching embeddings in the representation space.

\begin{figure}[!ht]
    \centering
    \includegraphics[width=\linewidth]{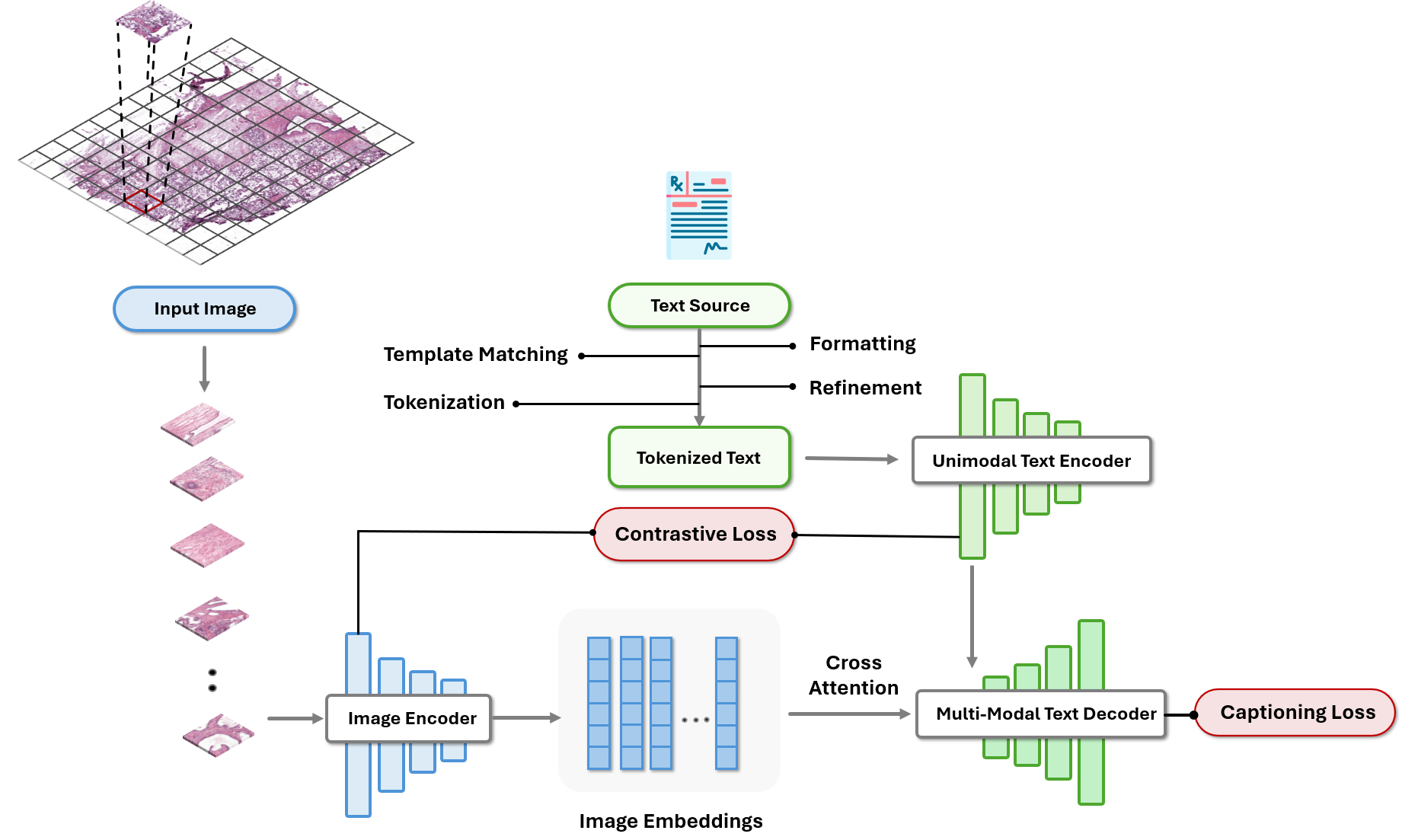}
    \caption{Visualization of the CoCa approach for the vision-language pre-training scheme.}
    \label{fig:coca}
\end{figure}

The second widely used approach is the contrastive captioner~(CoCa)~\cite{yu2022coca} framework shown in Fig.~\ref{fig:coca}. 
It contains three different modules; an image encoder, an unimodal text encoder and a multi-modal text decoder. It is trained by leveraging both captioning loss which follows a generative objective and contrastive loss which follows a contrastive objective. The input image patches are passed through an image encoder to generate image embeddings which is used to provide cross-attention to the multi-modal text decoder. The combination of generative and contrastive objectives forces the model to learn rich vision-language representation space.

Details of vision modules, language modules, additional modules and specifics of the pre-training process of FMs are outlined in Table~\ref{tab:vision_language_pretraing}.

One recent work KEP~\cite{zhou2024knowledge} proposes an entirely new approach in visual-language SSPT named knowledge enhanced pre-training by utilizing PathKT, a pathology knowledge tree curated by the same research.  This is different from the CLIP and CoCa pre-training techniques as it uses a novel knowledge encoder and knowledge distillation technique.

\subsection{Instruction-Tuning Phase}\label{sec:ins_tune}

For CPath, the instruction-tuning phase provides the model with conversational ability i.e. a user can prompt the model and the model will respond according to the prompt. Visual-instruction tuning adds the ability to provide an image in addition to natural language user prompts. Among the surveyed papers in FMs, PathChat and PathAsst perform instruction-tuning with self-curated datasets PathChatInstruct and PathInstruct, respectively. Both works adopt the strategy employed by LLaVA and LLaVA-1.5 which are the pioneering work in visual instruction tuning. A summary of the instruction tuning process is provided in Table~\ref{tab:vision_language_pretraing}. For both PathChat and PathAsst, the instruction tuning phase is subdivided into two phases. For both models in the first phase, the parameters of the vision encoder are frozen and the layer/module connected to the vision encoder (for PathChat it is a multimodal projector, for PathAsst it is fully connected layers) is trained. In the second phase, the model is fine-tuned with instruction-following data. More details about visual instruction tuning and a description of research work exclusively focusing on this aspect in CPath are provided in section~\ref{sec:v_tune}.

\subsection{Downstream Tasks and Datasets}\label{downstream}
FMs have the ability to adapt to a vast array of tasks by utilizing the representation space learned during SSPT.

\begin{table}[!h]
\caption{Different types of downstream tasks and corresponding research works}
\centering
\renewcommand{\arraystretch}{1.2}
\setlength{\tabcolsep}{3pt}
\resizebox{\linewidth}{!}{
\begin{tabular}{l|lc}
\bottomrule
\textbf{Downstream tasks} & \textbf{\makecell[l]{Performed By}} \\ \hline

 \makecell[l]{Disease/Cancer/Tissue/Tumor/Molecular SubTyping}             &  
   \makecell[l]{ \cite{huang2023visual}, \cite{lu2024visual},\cite{shaikovski2024prism},\cite{xu2024whole},\cite{hua2024pathoduet},\cite{chen2024towards},\\ \cite{yang2024foundation},\cite{dippel2024rudolfv},\cite{zhou2024knowledge},\cite{wang2022transformer},\cite{juyal2024pluto}} \\

    Cancer Detection             &       \cite{zimmermann2024virchow},\cite{huang2023visual},\cite{shaikovski2024prism},\cite{hua2024pathoduet},\cite{campanella2024computational},\cite{yang2024foundation},\cite{vorontsov2024foundation},\cite{wang2022transformer} \\

    Tumor Detection             &       \cite{huang2023visual}, \cite{sun2024pathasst},\cite{chen2024towards} \\

    Cancer Grading  &  \cite{lu2024visual},\cite{chen2024towards}\\

     Image/Tissue/Tumor/Gland/Nuclei Segmentation             &  
    \cite{lu2024visual},\cite{chen2024towards},\cite{dippel2024rudolfv}, \cite{wang2022transformer},\cite{juyal2024pluto} \\

      Survival Prediction             &  \cite{yang2024foundation}
    \\
      Text-to-Image Retrieval             &     \cite{huang2023visual},\cite{lu2024visual},\cite{zhou2024knowledge}\\

    Image-to-Text Retrieval             &     \cite{lu2024visual},\cite{zhou2024knowledge}  \\

       Image-to-Image Retrieval             &     \cite{huang2023visual},\cite{chen2024towards},\cite{dippel2024rudolfv},\cite{ma2024towards}   \\

         Image Captioning             &     \cite{lu2024visual}  \\

            Pattern/Tissue/Image Classification             &     \cite{lu2024visual},\cite{sun2024pathasst},\cite{chen2024towards}  \\
 Biomarker Prediction/Detection/Screening/Scoring            &     \cite{shaikovski2024prism},\cite{chen2024towards},\cite{campanella2024computational},\cite{dippel2024rudolfv},\cite{vorontsov2024foundation}  \\
  Metastasis Detection            &     \cite{chen2024towards},\cite{filiot2023scaling}   \\

    Organ Transplant Assessment            &     \cite{chen2024towards}   \\

  Mutation Detection/Prediction            &     \cite{xu2024whole},\cite{chen2024towards},\cite{campanella2024computational},\cite{filiot2023scaling}  \\
           VQA             &     \cite{sun2024pathasst},\cite{lu2024multimodal},\cite{ma2024towards}   \\

             Report Generation            &     \cite{shaikovski2024prism},\cite{ma2024towards}   \\

             Survival Analysis & \cite{zimmermann2024virchow}, \cite{ma2024towards},\cite{filiot2023scaling}\\ 

            Conversational Agent            &     \cite{sun2024pathasst},\cite{lu2024multimodal}  \\
\toprule
\end{tabular}}
\label{tab:downstream_tasks}
\end{table}
Note that, in the pre-training phase FMs were never trained for any of these specific tasks. At the end of the pre-training of FMs, one of the following strategies is adopted to perform a specific task.

\begin{enumerate}
    \item \textbf{Linear Probing:} This is a commonly used technique where a linear classifier/regressor is trained on top of the pre-trained model. During the training of the linear layers, the parameters of the pre-trained model are kept frozen. Depending on the specifics of the tasks, the corresponding loss function and update rule are determined. This is a computationally cheap way to adapt to a downstream task as the parameters of the pre-trained model do not need to be updated.

    \item \textbf{KNN Probing:} This is yet another approach to adopt a FM for a specific downstream task by utilizing K-nearest neighbors algorithm. In KNN probing a specific representation  from learned  representation space (through SSPT) is used to find its nearest neighbors.

\begin{figure}[!h]
    \centering
    \includegraphics[width=0.85\linewidth]{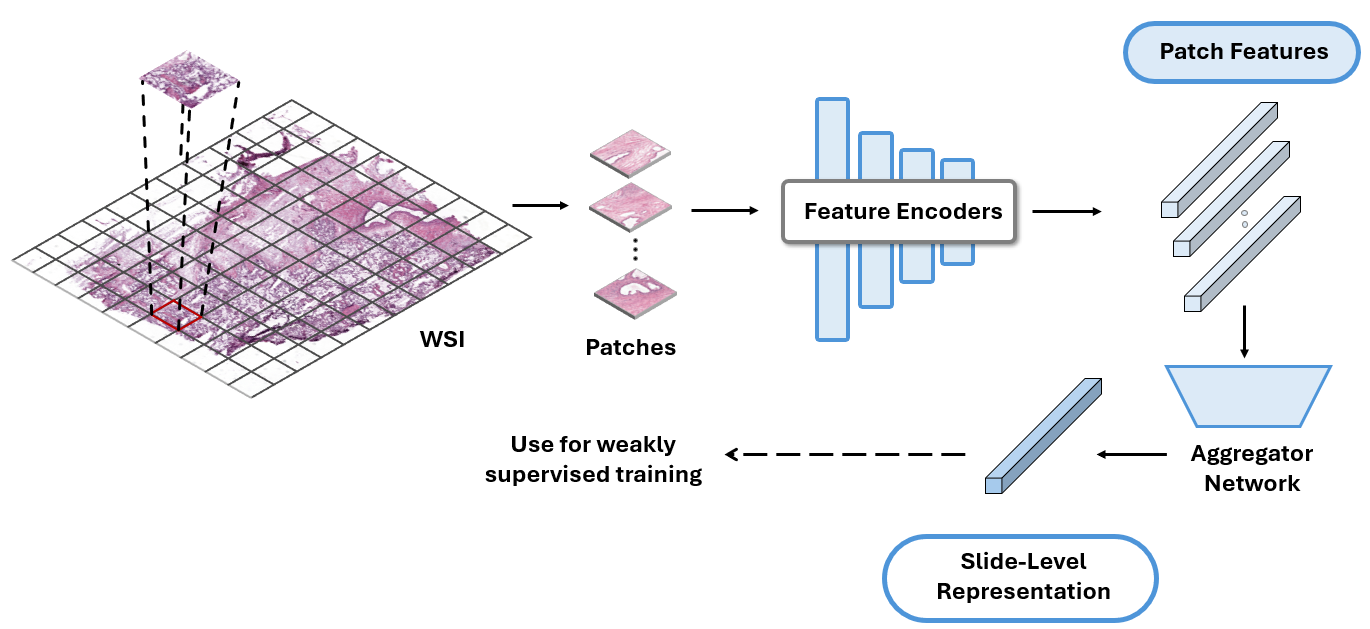}
    \caption{Aggregate patch-level features through an aggregator network to get a slide-level representation and use it for weakly supervised training.}
    \label{fig:agg}
\end{figure}

    \item  \textbf{Fine-Tuning:} This is similar to linear probing as a classifier/regressor is added on top of the pre-trained model, but the major difference is the parameter of the pre-trained models is also updated during fine-tuning. Hence, it is computationally much more costly compared to linear probing. This is sometimes also referred to as the supervised training phase.  Note that, most of the time annotation is only available on slide-level. Hence, if the slide-level label is utilized it is called weakly supervised training. To attain the slide-level representation from the trained model (which is trained to generate patch-level embeddings), typically a aggregator model is used. A visualization of attaining the slide level representation from the patch embeddings is shown in Fig.~\ref{fig:agg}.
    
\end{enumerate}

In Table~\ref{tab:downstream_tasks}, a summary of downstream tasks is provided along with research works performing these tasks. In Table~\ref{tab:dt_datasets_list} a list of downstream datasets utilized by different FMs is provided.

\begin{figure}[!ht]
    \centering
    \includegraphics[width=0.75\linewidth]{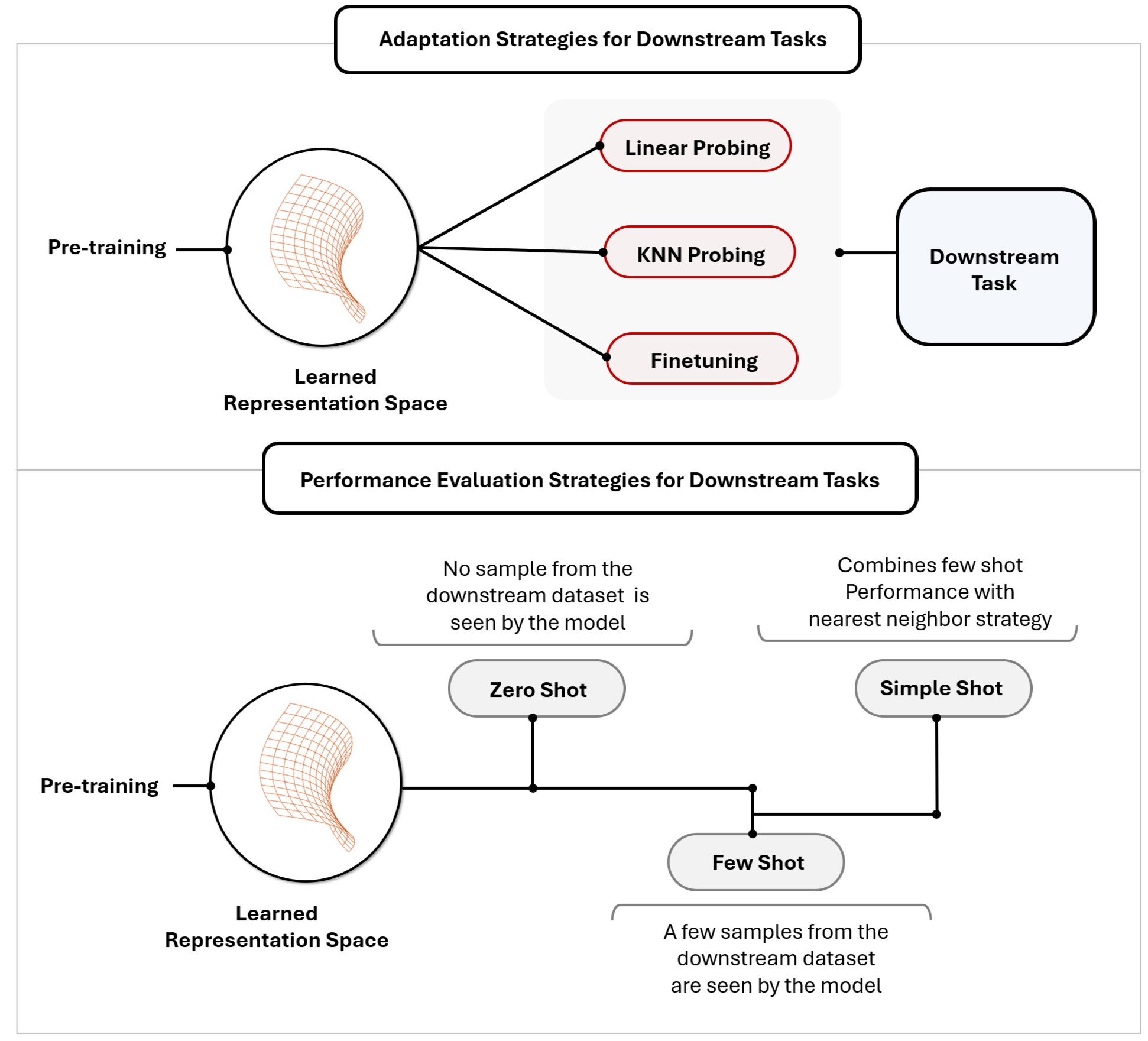}
    \caption{Adaptation and performance evaluation strategy for downstream tasks.}
    \label{fig:adaptation_dt}
\end{figure}

Another aspect to consider is how the performance evaluation is conducted. Among the surveyed articles, there are three different strategies that are employed.

\begin{enumerate}
    \item \textbf{Zero Shot Evaluation:} In zero shot evaluation, the pre-trained model is directly used in a downstream task without probing or fine-tuning the pre-trained model with any samples of the downstream dataset. This provides a direct assessment of learned representation in the pre-training phase i.e. evaluates the quality of the generated embeddings from the pre-trained model.  This is the most common approach among the surveyed articles.

    \item \textbf{Few Shot Evaluation:} In few shot evaluation, the pre-trained model sees only a few examples from the downstream task dataset. This is sometimes mentioned as K-shot evaluation where K is the number of data points seen by the model.

    \item \textbf{Simple Shot Evaluation:} Simpleshot~\cite{wang2019simpleshot} is a unique variation of the few shot evaluation methods which is only been explored in UNI. It combines  few shot learning with a nearest-neighbor classifier.
    
\end{enumerate}

A summary visualization for these adaptation and evaluation strategies is given in Fig.~\ref{fig:adaptation_dt}.

\subsection{Framework, Benchmarking  and Adaptation of FMs}

There are several research works in CPath that do not directly propose a FM but introduce the framework of FMs, provide benchmarks and comparisons between FMs and adapt the FMs for efficient training. 
\begin{table}
\caption{Evaluation datasets and the associated models utilizing the dataset along with the data source}
\centering
\renewcommand{\arraystretch}{1.2}
\setlength{\tabcolsep}{4pt}
\resizebox{0.9\linewidth}{!}{
\begin{tabular}{l|lc}
\bottomrule
\textbf{Dataset} & \textbf{\makecell[l]{Used By}} &  \makecell[c]{\textbf{Source}} \\ \hline
    
    PanNuke &      \cite{huang2023visual},\cite{juyal2024pluto}   & \href{https://warwick.ac.uk/fac/cross_fac/tia/data/pannuke}{\ding{51}}  \\

   HEST-1K &     \cite{zimmermann2024virchow}  & \href{https://github.com/mahmoodlab/hest}{\ding{51}}  \\

 DLBCL-Morph & \cite{zimmermann2024virchow} & \href{https://www.cancerimagingarchive.net/collection/dlbcl-morphology/}{\ding{51}}\\
      TissueNet &      \cite{wang2022transformer}   & \href{https://www.drivendata.org/competitions/67/competition-cervical-biopsy/page/254/}{\ding{51}}  \\

    GlaS & \cite{juyal2024pluto} & \href{https://www.kaggle.com/datasets/sani84/glasmiccai2015-gland-segmentation}{\ding{51}}\\

   CRAG & \cite{wang2022transformer} & \href{https://github.com/XiaoyuZHK/CRAG-Dataset_Aug_ToCOCO}{\ding{51}}\\
    
    RenalCell&     \cite{zhou2024knowledge}        &  \href{https://github.com/vahvero/RCC_textures_and_lymphocytes_publication_image_analysis}{\ding{51}}  \\
SkinCancer & \cite{zhou2024knowledge}                    & \href{https://heidata.uni-heidelberg.de/dataset.xhtml?persistentId=doi:10.11588/data/7QCR8S}{\ding{51}}     \\

UBC-OCEAN & \cite{ma2024towards}                 & \href{https://www.kaggle.com/competitions/UBC-OCEAN/data}{\ding{51}}     \\

Chaoyang & \cite{ma2024towards}                 & \href{https://bupt-ai-cz.github.io/HSA-NRL/}{\ding{51}}     \\

    DigestPath & \cite{huang2023visual}, \cite{lu2024visual}      & \href{https://digestpath2019.grand-challenge.org/}{\ding{51}}   \\
    WSSS4LUAD               &    \cite{huang2023visual},\cite{sun2024pathasst},\cite{lu2024visual},\cite{zhou2024knowledge},\cite{ma2024towards}     &   \href{https://wsss4luad.grand-challenge.org/}{\ding{51}} \\

    PatchCamelyon (PCam) &            \cite{zimmermann2024virchow},\cite{nechaev2024hibou},\cite{dippel2024rudolfv},\cite{vorontsov2024foundation},\cite{ma2024towards}&   \href{https://patchcamelyon.grand-challenge.org/Download/}{\ding{51}} \\
    MHIST&        \cite{zimmermann2024virchow},\cite{nechaev2024hibou},\cite{dippel2024rudolfv},\cite{vorontsov2024foundation}&  \href{https://bmirds.github.io/MHIST/}{\ding{51}}  \\

    BACH&       \cite{yang2024foundation},\cite{chen2024towards},\cite{zhou2024knowledge},\cite{ma2024towards}    &  \href{https://iciar2018-challenge.grand-challenge.org/Dataset/}{\ding{51}}  \\
    SICAP, SICAPv2&  \cite{lu2024visual}, \cite{zhou2024knowledge}&  \href{https://data.mendeley.com/datasets/9xxm58dvs3/2}{\ding{51}}  \\

     CoNSeP&     \cite{vorontsov2024foundation}        &  \href{https://github.com/vqdang/hover_net}{\ding{51}}  \\
     
    KIMIA Path24C   &  \cite{huang2023visual}       &  \href{https://kimialab.uwaterloo.ca/kimia/index.php/pathology-images-kimia-path24/}{\ding{51}}  \\
 
     DHMC    &   \cite{lu2024visual},\cite{chen2024towards}     & \href{https://bmirds.github.io/LungCancer/}{\ding{51}}   \\

     EBRAINS    &   \cite{lu2024visual},\cite{chen2024towards},\cite{ma2024towards}     & \href{https://search.kg.ebrains.eu/instances/Dataset/8fc108ab-e2b4-406f-8999-60269dc1f994}{\ding{51}}   \\

    AGGC  &   \cite{lu2024visual},\cite{chen2024towards}     & \href{https://aggc22.grand-challenge.org/Data/}{\ding{51}}   \\

     PANDA   &   \cite{lu2024visual},\cite{chen2024towards},\cite{ma2024towards}     & \href{https://panda.grand-challenge.org/data/}{\ding{51}}   \\

     MSK-IMPACT   &   \cite{shaikovski2024prism},\cite{vorontsov2024foundation}     & \href{https://www.mskcc.org/msk-impact}{\ding{51}}   \\

     SegPath  &   \cite{chen2024towards}     & \href{https://dakomura.github.io/SegPath/}{\ding{51}}   \\ 

    BRACS &   \cite{chen2024towards},\cite{ma2024towards}     & \href{https://www.bracs.icar.cnr.it/}{\ding{51}}   \\ 

    UniToPatho   &   \cite{chen2024towards},\cite{ma2024towards},\cite{wang2022transformer}     & \href{https://zenodo.org/records/4643645}{\ding{51}}   \\ 

     HunCRC  &   \cite{chen2024towards}     & \href{https://www.cancerimagingarchive.net/collection/hungarian-colorectal-screening/}{\ding{51}}   \\ 

       BreakHis&       \cite{yang2024foundation},\cite{ma2024towards}    & \href{https://web.inf.ufpr.br/vri/databases/breast-cancer-histopathological-database-breakhis/}{\ding{51}}   \\

       MSI-CRC \& MSI-STAD&    \cite{nechaev2024hibou},\cite{dippel2024rudolfv}    & \href{https://zenodo.org/records/2530835}{\ding{51}}   \\

       TIL-DET&    \cite{nechaev2024hibou},\cite{dippel2024rudolfv}    & \href{https://zenodo.org/records/6604094}{\ding{51}}   \\

          MIDOG&     \cite{zimmermann2024virchow}, \cite{vorontsov2024foundation},\cite{wang2022transformer}   & \href{https://midog.deepmicroscopy.org/download-dataset/}{\ding{51}}   \\
\hline

CAMELYON16   &    \cite{yang2024foundation},\cite{hua2024pathoduet},\cite{chen2024towards},\cite{ma2024towards},\cite{filiot2023scaling},\cite{wang2022transformer}    & \href{https://camelyon16.grand-challenge.org/}{\ding{51}}   \\
 
 CAMELYON17-WILDS  &   \cite{zimmermann2024virchow},\cite{chen2024towards},\cite{vorontsov2024foundation},\cite{juyal2024pluto},\cite{ma2024towards},\cite{filiot2023scaling}     & \href{https://wilds.stanford.edu/datasets/}{\ding{51}}   \\
\hline

\multirow{4}{*}{
\begin{forest}
  for tree={
    grow'=0,
    child anchor=west,
    parent anchor=south,
    anchor=west,
    calign=first,
    s sep+=-5pt,
    inner sep=3.5pt,
    edge path={
      \noexpand\path [draw, \forestoption{edge}]
      (!u.south west) +(5pt,0) |- (.child anchor)\forestoption{edge label};
    },
    before typesetting nodes={
      if n=1{
        insert before={[, phantom, my tier]},
      }{},
    },
    my tier,
    fit=band,
    before computing xy={
      l=30pt,
    },
  }
[Kather colon
    [CRC100K]
    [NCT-CRC-HE-100K]
    [NCT-CRC-HE-100K-NONORM]
]
\end{forest}

}       &   \cite{huang2023visual},\cite{yang2024foundation},\cite{hua2024pathoduet},\cite{juyal2024pluto}, \cite{zhou2024knowledge}     &   \multirow{4}{*}{\href{https://zenodo.org/records/1214456}{\ding{51}}}  \\
           &   \cite{sun2024pathasst},\cite{lu2024visual},\cite{chen2024towards},\cite{nechaev2024hibou},\cite{dippel2024rudolfv},\cite{ma2024towards}     &  \\
              &   \cite{yang2024foundation},\cite{vorontsov2024foundation},\cite{zhou2024knowledge}  &  \\

                &  \cite{zimmermann2024virchow}, \cite{vorontsov2024foundation}   &  \\
\hline
         
  \multirow{3}{*}{
\begin{forest}
  for tree={
    grow'=0,
    child anchor=west,
    parent anchor=south,
    anchor=west,
    calign=first,
    s sep+=-5pt,
    inner sep=3.5pt,
    edge path={
      \noexpand\path [draw, \forestoption{edge}]
      (!u.south west) +(5pt,0) |- (.child anchor)\forestoption{edge label};
    },
    before typesetting nodes={
      if n=1{
        insert before={[, phantom, my tier]},
      }{},
    },
    my tier,
    fit=band,
    before computing xy={
      l=30pt,
    },
  }
[LC25000
    [LC25000Colon]
    [LC25000Lung]
]
\end{forest}}&      \cite{sun2024pathasst},\cite{yang2024foundation},\cite{zhou2024knowledge}       &  \multirow{3}{*}{\href{https://github.com/tampapath/lung_colon_image_set}{\ding{51}}} \\

      &   \cite{sun2024pathasst},\cite{ikezogwo2024quilt}       &    \\
       &      \cite{sun2024pathasst},\cite{ikezogwo2024quilt}    &    \\

\hline
  TCGA Uniform&       \cite{chen2024towards}     & \href{https://zenodo.org/records/5889558}{\ding{51}}   \\ 

    TCGA CRC-MSI&       \cite{chen2024towards},\cite{ma2024towards},\cite{vorontsov2024foundation}     & \href{https://zenodo.org/records/3832231}{\ding{51}}   \\
    
    TCGA-TILs&       \cite{zimmermann2024virchow}, \cite{vorontsov2024foundation},\cite{chen2024towards}     & \href{https://zenodo.org/records/6604094}{\ding{51}}   \\ 

\hline

\multirow{4}{*}{
\begin{forest}
  for tree={
    grow'=0,
    child anchor=west,
    parent anchor=south,
    anchor=west,
    calign=first,
    s sep+=-5pt,
    inner sep=3.5pt,
    edge path={
      \noexpand\path [draw, \forestoption{edge}]
      (!u.south west) +(5pt,0) |- (.child anchor)\forestoption{edge label};
    },
    before typesetting nodes={
      if n=1{
        insert before={[, phantom, my tier]},
      }{},
    },
    my tier,
    fit=band,
    before computing xy={
      l=30pt,
    },
  }
[TCGA
    [TCGA BRCA]
    [TCGA RCC]
    [TCGA NSCLC]
]
\end{forest}}&   \cite{filiot2023scaling}          &  \multirow{4}{*}{\href{https://portal.gdc.cancer.gov/}{\ding{51}}} \\
       &  \cite{lu2024visual},\cite{shaikovski2024prism},\cite{yang2024foundation},\cite{nechaev2024hibou},\cite{zhou2024knowledge}       &    \\
        &      \cite{lu2024visual},\cite{yang2024foundation},\cite{nechaev2024hibou},\cite{hua2024pathoduet},\cite{zhou2024knowledge},\cite{wang2022transformer}   &    \\
     &         \cite{lu2024visual},\cite{shaikovski2024prism},\cite{yang2024foundation},\cite{nechaev2024hibou},\cite{hua2024pathoduet},\cite{juyal2024pluto},\cite{zhou2024knowledge},\cite{wang2022transformer}  &    \\

\bottomrule
\multicolumn{3}{l}{ \makecell[l]{\textit{Mentioned data sources are shared through platforms like Zenodo, GitHub,} \\ \textit{official challenge websites. TCGA dataset needs to be downloaded from the official portal.}}}

\end{tabular}}
\label{tab:dt_datasets_list}
\end{table}

\subsubsection{Frameworks}  eva~\cite{aben2024towards,gatopoulos2024eva} is a framework for VFMs in CPath which abstracts a lot of complexity of VFMs. In addition, it facilitates the reproducibility of VFMs for fair comparison and provides an interface to evaluate publicly available downstream datasets.

\subsubsection{Benchmarking}

Another category is benchmark analysis of FMs \cite{lu2024multiple, mallya2024benchmarking, campanella2024clinical, zheng2024benchmarking,aswolinskiy2024impact,alfasly2024foundation,lai2023domain} in CPath. The work in~\cite{lu2024multiple} benchmarks the performance of $6$ foundation models (CTransPath, PathoDuet, PLIP, CONCH and UNI) across $5$ clinically relevant prediction tasks. The contribution of~\cite{mallya2024benchmarking}  is benchmarking  FMs which include CTransPath, Lunit, Phikon, PLIP, UNI and Virchow for ovarian cancer bevacizumab  reatment response prediction. Another work~\cite{campanella2024clinical} analyzes the performance of FMs on a large and diverse data set collected from two medical centers. The FMs (CTransPath, UNI, Virchow and Prov-GigaPath) were benchmarked on $3$ broad downstream tasks which include disease detection, biomarker prediction and treatment outcome prediction. In~\cite{zheng2024benchmarking}, the authors benchmarks PathCLP~\cite{sun2024pathasst} on various corrupted images (with $7$ corruption types) for $2$ downstream datasets. The authors of~\cite{aswolinskiy2024impact} investigate how layer selection for FMs (CTransPath, Phikon, Lunit + other encoders) affects the downstream task performance. The work done in \cite{alfasly2024foundation} analyzes different FMs and evaluates their performance on $8$ datasets. In~\cite{lai2023domain}, the authors evaluates the performance of histopathology-specific SSL methods for $17$ unique tissue types and $12$ unique cancer types.

\subsubsection{Adaptation of FMs}
Another category is the adaptation of existing FMs to carry out tasks such as low-resource fine-tuning\cite{roth2024low} and multi-modal prompt-tuning\cite{lu2024pathotune}. The work carried out in~\cite{roth2024low}  fine-tunes a FM with a single GPU and shows that it can outperform SOTA feature extractors. PathoTune~\cite{lu2024pathotune} adapts a visual or
pathology-specific FM to downstream tasks using multi-modal prompts. In~\cite{yin2024prompting} the authors proposes a task-specific visual prompting approach to tune VFMs. The work done in~\cite{ferber2024context} utilizes in-context learning to learn from prompts without parameter updates. They specifically utilizes GPT-4 with vision capability and performs evaluation on $3$ downstream datasets.

Other than the surveyed FMs, there are more recent works of FM in CPath like mSTAR~\cite{xu2024multimodal}. However, as it includes RNA-Seq data in addition to pathology reports it falls outside the scope of this paper. Another foundation model we do not include is H-optimus-0~\cite{hoptimus0} developed by  Bioptimus as it is not associated with a publication.

\begin{table*}[p]
\centering
\caption{Summary of Vision-Language Models in Computational Pathology}
\renewcommand{\arraystretch}{1.2}
\setlength{\tabcolsep}{10pt}
\resizebox{0.98\linewidth}{!}{
\begin{tabular}{lP{10cm}P{5cm}P{8cm}c}
\toprule
\textbf{Model}                 & \makecell[c]{\textbf{Architecture and Utilized Models/Frameworks}} & \textbf{Dataset} & \makecell[c]{\textbf{Contribution}} & \textbf{Availability} \\ \hline

TraP-VQA~\cite{naseem2022vision} & \begin{itemize}[left=0pt]
    \item 
Question feature extraction with  BioELMo~\cite{jin2019probing} and BiLSTM \item  Image feature extraction with ResNet-50 and CNN layers \item  Transformer encoder to fuse question and image feature, transformer decoder to upsample fused features         \end{itemize}           & PathVQA              &  \begin{itemize}[left=0pt]
    \item Performs VQA \item 
Provides interpretability in text domain with SHAP~\cite{lundberg2017unified} \item Provides interpretability in image domain with GradCAM~\cite{selvaraju2017grad}     \end{itemize}        & \textbf{---}                  \\ \midrule

 FSWC~\cite{qu2024rise} & \begin{itemize}[left=0pt]
    \item 
 CLIP for image and text feature extraction \item GPT-4 to generate instance-level and slide-level prompt groups which introduce pathological prior knowledge into the model \end{itemize}            & \begin{itemize}[left=0pt]
     \item 
 Camelyon 16  \item  TCGA-Lung  \item In-house cervical cancer dataset  \end{itemize}           & \begin{itemize}[left=0pt]
  \item Performs few-shot weakly supervised WSI classification
    \item 
Proposes a two-level (instance-level and slide-level)
prompt learning MIL framework named \textbf{TOP} \item Introduces a prompt guided instance pooling to generate slide-level feature   \end{itemize}                & \href{https://github.com/miccaiif/TOP}{\ding{51}}    \\ \midrule

ViLA-MIL~\cite{shi2024vila}  &  \begin{itemize}[left=0pt]

\item GPT-3.5  as
the frozen LLM which generates visual descriptive text for WSI at 5x and 10x resolution  based on class level question prompt \item ResNet-50 
as the image encoder and corresponding CLIP
transformer is used as the text encoder
\item  Prototype-guided patch decoder which generates slide features and context-guided
text decoder which generates text-features 
\end{itemize} & \begin{itemize}[left=0pt]
    \item TIHD-RCC (in-house dataset) \item  TCGA-RCC 
\item TCGA-Lung
\end{itemize}& \begin{itemize}[left=0pt]
    \item Performs WSI classification with a MIL framework that utilizes information from 5x and 10x. In addition, incorporates information from  LLM generated visual descriptive text for both 5x and 10x scale WSIs
    
    \item Introduces a novel  prototype-guided patch decoder
that progressively aggregates the patch features 
    \item Introduces a context-guided text decoder to refine the text
prompt features by leveraging multi-granular image contexts

\end{itemize}& \href{https://github.com/Jiangbo-Shi/ViLa-MIL}{\ding{51}}\\ \midrule

PathM3~\cite{zhou2024pathm3}& \begin{itemize}[left=0pt]

    \item ViT-G as the image encoder
    \item Query-based transformer to fuse image embeddings of WSIs and corresponding captions
    \item Frozen LLM FlanT5~\cite{chung2024scaling} for caption generation
\end{itemize}& PatchGastric~\cite{tsuneki2022inference} & \begin{itemize}[left=0pt]
    \item  Performs WSI classification and captioning through a multi-modal, multi-task, multi-instance learning framework  
    \item Leverages limited WSI captions during training \item Develops multi-task joint learning inspired from \cite{wang2018tienet}
\end{itemize} & \textbf{---} \\ \midrule

HLSS~\cite{watawana2024hierarchical} & \begin{itemize}[left=0pt]
    \item  ResNet-50  as  visual encoder with CLIP pre-training 
    \item  Language encoder from CLIP
    \item  A positive pairing module (PPM) which consists of three parallel reshape layer followed by MLP
    \item A cross-modal alignment
(CAM) module which computes cosine similarity
    
\end{itemize}& \begin{itemize}[left=0pt]
    \item OpenSRH~\cite{jiang2022opensrh} \item TCGA
\end{itemize} & \begin{itemize}[left=0pt]
    \item Performs  hierarchical (patient-slide-patch hierarchy) text-to-vision alignment for SSL
    \item  Proposes a positive pairing module
(PPM) and  a cross-modal alignment module (CAM)
\end{itemize} & \href{https://github.com/Hasindri/HLSS} {\ding{51}} \\ \midrule

FiVE~\cite{li2024generalizable} & \begin{itemize}[left=0pt]

    \item  ResNet as image encoder following~\cite{li2021dual} 
    \item Pre-trained BioClinicalBERT~\cite{wolf2019huggingface} as text encoder and utilizing  LoRA~\cite{hu2021lora} for fine-tuning 
    \item GPT-4 to generate fine-grained pathological descriptions based on raw pathology reports and expert inquiries
    \item An instance aggregator module consisting of a self-attention module and a cross-attention module that fuses image instance embeddings and prompt embeddings to create bag-level feature 
\end{itemize} & \begin{itemize}[left=0pt]
    \item Camelyon 16 \item TCGA-Lung
\end{itemize} & \begin{itemize}[left=0pt]
    \item Incorporates a patch-samnpling strategy for optimize training efficiency of the model
    \item Proposes a task-specific fine-grained semantics (TFS) module

    \item Performs zero shot histological subtype classification, few shot classification and supervised classification with pre-training  
\end{itemize} & \href{https://github.com/ls1rius/wsi_five}{\ding{51}} \\ \midrule

CPLIP~\cite{javed2024cplip} & \begin{itemize}[left=0pt]

    \item MI-Zero~\cite{lu2023visual} to identity-related prompts based on an image  by utilizing similarity metric
    \item GPT-3 to categorize and transform prompts into five variations
    \item PILP to match the transformed prompts with corresponding images from OpenPath
\end{itemize}& \begin{itemize}[left=0pt]
    \item  CRC100K \item WSSS4LUAD \item PanNuke \item DigestPath \item SICAP \item  Camelyon 16 \item  TCGA-BRCA \item TCGA-RCC \item TCGA-NSCLC
\end{itemize} & \begin{itemize}[left=0pt]
    \item  Creates  a pathology-specific dictionary using a range
of publicly available online glossaries

    \item Introduces a many-to-many contrastive learning instead of traditional one-to-one contrastive learning 
\end{itemize}&  \href{https://github.com/iyyakuttiiyappan/CPLIP}{\ding{51}}
\\ \midrule

 HistGen~\cite{guo2024histgen} & \begin{itemize}[left=0pt]
     \item GPT-4 to clean and summarize reports curated from TCGA
     \item A ViT-L model is pre-trained with 200 million patches by utilizing DINOv2 and used as feature extractor
 \end{itemize} & \begin{itemize}[left=0pt]
     \item UBC-OCEAN~\cite{asadi2024machine,farahani2022deep}
     \item TUPAC16~\cite{veta2019predicting}
     \item Camelyon 16 and 17
     \item TCGA-BRCA
     \item TCGA-STAD
     \item TCGA-KIRC
     \item TCGA-KIRP
     \item TCGA-LUAD
     \item TCGA-COADREAD
 \end{itemize} & \begin{itemize}[left=0pt]

     \item Performs WSI report generation, cancer subtyping and survival analysis 
     \item Develops a local-global hierarchical encoder module to capture features at different levels (region-to-slide) 

     \item Develops a cross-modal context-aware learning module to align and ensure interaction between different modalities 
 \end{itemize}& \href{https://github.com/dddavid4real/HistGen}{\ding{51}}
 \\ 
 \midrule

CLIPath~\cite{lai2023clipath} & \begin{itemize}[left=0pt]
    \item ResNet-50 as the vision encoder in CLIP
    \item $4$ fully-connected layer after the vision encoder 
    \item Transformer as the text encoder in CLIP
\end{itemize} & \begin{itemize}[left=0pt]
    \item  PatchCamelyon (PCam) \item MHIST \end{itemize}& \begin{itemize}[left=0pt]
    \item Develops residual feature connection (RFC) to fine-tune CLIP
with a small amount of trainable parameters and this also fuses the existing knowledge from pre-trained CLIP and the newly learned knowledge related to pathology-specific tasks 
\end{itemize} & \textbf{---} \\ \midrule
MI-Zero~\cite{lu2023visual} & \begin{itemize}[left=0pt]
    \item HistPathGPT: Unimodal pre-training of GPT-style transformer (same architecture as GPT 2-medium~\cite{radford2019language}) which is used as text encoder.
    \item Additionally, BioClinicalBERT~\cite{alsentzer2019publicly}  and  PubMedBERT~\cite{gu2021domain} are considered as text encoders.

    \item Existing encoders like CTransPath~\cite{wang2022transformer} (which is based on Swin Transformer~\cite{liu2021swin}) or ViT-S (ImageNet initialization or pre-trained with MoCov3)
\end{itemize} & \begin{itemize}[left=0pt]
    \item 3 WSI datasets
from Brigham and Women’s Hospital (in-house dataset) named Independent BRCA, Independent NSCLC, Independent RCC

 \item TCGA-BRCA
  \item TCGA-NSCLC
  \item TCGA-RCC
\end{itemize} & \begin{itemize}[left=0pt]
    \item Performs cancer subtyping with a VLM
    \item Develops a custom pathology domain-specific pre-trained text encoder called HistPathGPT 
    \item Introduces 33,480 image-caption pair dataset 
\end{itemize} & \href{https://github.com/mahmoodlab/MI-Zero}{\ding{51}}
\\ \midrule
HistoGPT~\cite{tran2024generating} & \begin{itemize}[left=0pt]
    \item The vision module is based on CTransPath

    \item BioGPT as the language module
    \item Image features sampled (with Perceiver Resampler~\cite{jaegle2021perceiver}) from the vision modules are integrated into the language module via interleaved gated cross-attention (XATTN) blocks~\cite{alayrac2022flamingo}

\end{itemize} & \begin{itemize}[left=0pt]
    \item  In-house dataset with 2 cohorts. Munich cohort with l 6,000 histology samples and M\"unster
cohort with 1,300 histology samples with all samples stained with H\&E \end{itemize}& \begin{itemize}[left=0pt]
    \item Performs histopathology report generation that provides a description of WSIs with high fidelity 

    \item Provides interpretability map which shows which word in the generated report corresponds to which region in a WSI
\end{itemize} & \href{https://github.com/marrlab/HistoGPT}{\ding{51}} \\ 
\bottomrule

\end{tabular}}
\label{tab:vlms}
\end{table*}

\begin{table*}[p]
\centering
\renewcommand{\arraystretch}{1.2}
\setlength{\tabcolsep}{10pt}
\resizebox{0.95\linewidth}{!}{
\begin{tabular}{lP{10cm}P{5cm}P{8cm}c}
\toprule
\textbf{Model}                 & \makecell[c]{\textbf{Architecture and Utilized Models/Frameworks}} & \textbf{Dataset} & \makecell[c]{\textbf{Contribution}} & \textbf{Availability} \\ \hline

PathAlign~\cite{ahmed2024pathalign} & \begin{itemize}[left=0pt]
    \item  PathSSL patch encoder which is based on ViT-S architecture and pre-trained following the approach in~\cite{lai2023domain} with masked
siamese networks~\cite{assran2022masked} SSL scheme  \item Q-Former from BLIP-2~\cite{li2023blip} as WSI-encoder \item Among two variants PathAlign-R and PathAlign-G, PathAlign-G uses  PaLM-2 S~\cite{anil2023palm} as the frozen LLM  \end{itemize}           & \begin{itemize}[left=0pt] \item In-house de-identified dataset (DS1) of WSIs paired with reports collected from a teaching hospital. The stain type for the WSIs are H\&E and IHC  \item TCGA  \end{itemize}       &  \begin{itemize}[left=0pt]
    \item  Performs WSI classification, image-to-text retrieval and text generation  \end{itemize}        & \textbf{---}                  \\ \midrule

MI-Gen~\cite{chen2023mi} & \begin{itemize}[left=0pt]
    \item ResNet and ViT (ImageNet initialization or hierarchical SSPT with HIPT~\cite{chen2022scaling})  as visual encoder 
    \item CNN layer in the hierarchical position-aware module (PAM) 

    \item Transformer as encoder-decoder (both vanilla transformer and Mem-Transformer~\cite{chen2020generating}). Additionally, CNN-RNN~\cite{vinyals2015show} and att-LSTM~\cite{xu2015show} is used for comparison
\end{itemize} & PathText dataset which was created using and WSI and pathology report from TCGA-BRCA & \begin{itemize}[left=0pt]

    \item Performs pathology report generation from WSI
    \item Additionally performs cancer subtyping and biomarker prediction
     \item Develops a
hierarchical position-aware module (PAM) 
    \item Introduces PathText dataset which contains 9,009 WSI-text
pairs
\end{itemize} & \href{https://github.com/cpystan/Wsi-Caption}{\ding{51}}
\\ \midrule
W2T~\cite{chen2024wsi} & \begin{itemize}[left=0pt]
    \item ResNet-50, ViT-S pre-trained with DINO or scheme followed in HIPT  as visual encoders to extract patch features 
    \item Text Encoders: PubMedBERT and BioClinicalBERT

    \item Co-attention mapping in the decoder to align visual and text features

\end{itemize} & WSI-VQA  & \begin{itemize}[left=0pt]
    \item Curates a WSI VQA dataset called WSI-VQA with 977 WSIs and 
8,672 Q/A

\item Performs VQA with the proposed WSI-VQA dataset

\item Co-attention mapping between word embeddings and WSIs are used for interpretability
\end{itemize} & \href{https://github.com/cpystan/WSI-VQA}{\ding{51}} \\ \midrule
\makecell[l]{PathGen-CLIP~\cite{sun2024pathgen}} & \begin{itemize}[left=0pt]

    \item OpenCLIP framework is utilized to train a pathology-specific CLIP model 
    \item The vision encoder of LLaVA v1.5 is replaced with the trained pathology-specific CLIP model
    \item PathGen-CLIP is combined with Vicuna LLM 
\end{itemize} & \begin{itemize}[left=0pt] \item PathGen-1.6M (for pre-training)  \item PatchCamelyon (Pcam)  \item CRC-100K \item SICAPv2 \item BACH \item Osteo \item SkinCancer \item  LC25000 \item Camelyon 16 and 17
 \end{itemize}& \begin{itemize}[left=0pt]
    \item Curates a large-scale dataset PathGen-1.6M with 1.6 million image-caption pairs
    \item Develops a pathology-specific large multi-modal modal with the capability to adapt to various downstream task
\end{itemize} & \href{https://github.com/superjamessyx/PathGen-1.6M}{\ding{51}} \\ \midrule

Quilt-Net~\cite{ikezogwo2024quilt} & \begin{itemize}[left=0pt]
    \item CLIP model (based on OpenCLIP framework) 
    \item ViT-B as the image encoder
    \item GPT-2 and PubMedBERT as text encoder
    
\end{itemize} & \begin{itemize}[left=0pt]
    \item Quilt-1M (for pre-training)
    \item PatchCamelyon (PCam)
    \item NCT-CRC-HE-100K 
    \item SICAPv2 
    \item Databiox 
    \item BACH
    \item Osteo
    \item Renal Cell 
    \item SkinCancer
    \item MHIST
    \item LC25000

\end{itemize} & \begin{itemize}[left=0pt]

    \item Curates a large-scale dataset Quilt-1M
    \item Performs zero-shot, linear-probing task and also cross-modal retrieval task
    \item Provides cross-modal attention mask with Grad-CAM
\end{itemize}  & \href{https://github.com/wisdomikezogwo/quilt1m}{\ding{51}} \\ \midrule

Guevara et al.~\cite{guevara2023caption} & \begin{itemize}[left=0pt]
    \item CLIP model
    \item HIPT as image encoder
    \item A transformer encoder to project encoded image embedding to match the dimension of text embedding
    \item PubMedBERT as text encoder

    \item For the caption model pre-trained Bio GPT 2 is used 

    \item GPT-3.5-turbo~\cite{ouyang2022training} to clean and refine captions. Additionally, the captions were machine-translated from two languages to English using the approach in~\cite{tiedemann2020opus} 
\end{itemize} & Self-curated dataset of WSIs and captions. WSIs are of colon polyps and biopsies. The captions contain possible 5 diagnostic labels normal, hyperplasia, low-grade dysplasia, high-grade dysplasia, or adenocarcinoma  & \begin{itemize}[left=0pt]
    \item Performs WSI classification and caption generation through the use of weakly supervised transformer-based models
\end{itemize} & \textbf{---} \\ \midrule
Hu et al.~\cite{hu2024histopathology} & \begin{itemize}[left=0pt]

 \item ResNet-50 pre-trained with BYOL~\cite{grill2020bootstrap} SSL scheme as patch encoder
    \item Anchor-based module as WSI encoder based on~\cite{zheng2023kernel} which uses kernel attention transformer
    \item Prompt-based text encoder which utilizes self-attention structure in BERT~\cite{devlin2018bert}
\end{itemize} & \begin{itemize}[left=0pt]
    \item  GastricADC
    \item In-house gastric dataset containing 3598 WSIs and reports
dataset
\end{itemize} & \begin{itemize}[left=0pt]
    \item Performs cross-modal retrieval tasks which includes image-to-image retrieval, image-to-text retrieval, text-to-image retrieval and text-to-text retrieval 
    \item Proposes a histopathology language-image representation learning framework

    \item Develops a prompt-based text representation learning scheme 
\end{itemize} &  \href{https://github.com/hudingyi/FGCR}{\ding{51}} \\  \midrule

PEMP~\cite{qu2024pathology} & \begin{itemize}[left=0pt]
    \item CLIP as the backbone of the proposed model
    \item A self-attention layer 
    \item A attention pooling layer
\end{itemize} &  \begin{itemize}[left=0pt]
    \item In-house datasets
    \item TCGA-CESC 
\end{itemize} & \begin{itemize}[left=0pt]

   \item Performs survival analysis, metastasis detection and cancer subtype classification 
   
    \item  Proposes a mechanism to introduce vision and text prior knowledge in the designed prompts (both static and learnable) at both patch and slide levels
    
    \item Develops a self-attention layer called a lightweight messenger and an attention pooling layer called summary layer
\end{itemize} & \textbf{---}
\\ \midrule

\makecell[l]{HistoCap~\cite{sengupta2023automatic}\\ and HistoCapBERT} & \begin{itemize}[left=0pt]

    \item  ResNet-18 to encode thumbnail image
    \item  LSTM decoder according to~\cite{xu2015show} as caption generator
    \item  Pre-trained HIPT encoder for WSI encoding 
    \item Trainable attention layer after HIPT encoder
    \item In HistocapBERT the LSTM decoder is replaced by a BioclinicalBERT
\end{itemize} & Dataset from Genotype-Tissue Expression (GTEx) portal& \begin{itemize}[left=0pt]
    \item Performs caption generation given a thumbnail image and a WSI image
\end{itemize} & \href{https://github.com/ssen7/histo_cap_generation_2}{\ding{51}}
\\ \midrule

PathCap~\cite{zhang2020evaluating} & \begin{itemize}[left=0pt]
    \item ResNet-18 to encode thumbnail image 
    \item ResNet-18 to encode WSI patches
    \item Trainable attention layer after the ResNet-18 used to encode WSI patches
    \item  LSTM as caption generation module
\end{itemize} & Dataset from Genotype-Tissue Ex-
pression (GTEx) portal & \begin{itemize}[left=0pt]
    \item Performs caption generation of  histopathology images using multi-scale view (thumbnail of WSIs and WSIs) 
\end{itemize}
 & \href{https://github.com/zhangrenyuuchicago/PathCap}{\ding{51}}
\\ \midrule
Elbedwehy et al.~\cite{elbedwehy2024enhanced} & \begin{itemize}[left=0pt]
    \item Vision encoders include VGG~\cite{simonyan2014very}, ResNet, PVT~\cite{wang2022pvt}, Swin-Large~\cite{liu2021swin}, ConvNexT-Large~\cite{liu2022convnet}
    \item Language decoders and pre-trained word embedding models include LSTM, RNN, Bi-directional RNN,  BioLinkBERT-Large~\cite{yasunaga2022linkbert}
\end{itemize} & PatchGastric & \begin{itemize}[left=0pt]
    \item Performs caption generation of WSIs 
\end{itemize} & \textbf{---}
\\ \midrule
PromptBio~\cite{zhang2024prompting} & \begin{itemize}[left=0pt]
    \item PLIP as coarse-grained pathology instance classifier which filters out patches given a patch and a prompt (keeps patches associated with cancer-associated stroma)

       \item IBMIL~\cite{lin2023interventional} to encode patches followed by a fully-connected layer
       \item GPT-4 to generate pathology descriptions
       \item Series of transformer and MLP layer which performs biomarker prediction  
\end{itemize} & \begin{itemize}[left=0pt]
    \item TCGA-CRC
    \item The Clinical Proteomic Tumor Analysis Consortium
(CPTAC) CRC dataset
\end{itemize} & \begin{itemize}[left=0pt]
    \item Performs biomarker prediction 
 
\end{itemize} & \href{https://github.com/DeepMed-Lab-ECNU/PromptBio}{\ding{51}} 
\\ \midrule

\end{tabular}}
\end{table*}

\begin{table*}[!t]
\centering
\renewcommand{\arraystretch}{1.2}
\setlength{\tabcolsep}{10pt}
\resizebox{\linewidth}{!}{
\begin{tabular}{lP{10cm}P{5cm}P{8cm}c}
\toprule
\textbf{Model}                 & \makecell[c]{\textbf{Architecture and Utilized Models/Frameworks}} & \textbf{Dataset} & \makecell[c]{\textbf{Contribution}} & \textbf{Availability} \\ \hline

SGMT~\cite{qin2023whole} & \begin{itemize}[left=0pt]
    \item ResNet-18 (pre-trained on ImageNet) as patch encoder
    \item Transformer encoder to encode the patch features
    \item  Transformer decoder as caption generation module
\end{itemize} & PatchGastric & \begin{itemize}[left=0pt]
    \item Performs caption generation of WSIs
    \item Proposes a novel mechanism to use subtype prediction as a guiding mechanism for the caption generation task
    \item Develops a random sampling and voting strategy to select patches
\end{itemize} & --- \\ \midrule
Tsuneki et al.~\cite{tsuneki2022inference} & \begin{itemize}[left=0pt]
    \item EfficientNetB3~\cite{tan2019efficientnet} and DenseNet121~\cite{huang2017densely} (pre-trained on ImageNet) as image encoder

    \item Average and global pooling layer after the image encoders 

    \item RNN decoder layer for caption generation 
    
\end{itemize} & PatchGastric & \begin{itemize}[left=0pt]
    \item Performs caption generation of WSIs
    \item Curates a captioned dataset of 262,777 patches from 991 WSIs
\end{itemize}& \href{https://github.com/masatsuneki/histopathology-image-caption}{\ding{51}} \\ \bottomrule
\end{tabular}}
\end{table*}

\section{Vision-Language Models}\label{sec:vision_language_models}
In this section, the recent works in CPath with VLMs are outlined. The details about architecture, used datasets and contribution of individual research work are listed in Table~\ref{tab:vlms}.

First, different categorizations of VLMs are provided (section~\ref{sec:cat_vlm}) to give insight into how VLMs are utilized to solve pathology-specific tasks. Second, the common architectural components in VLMs and adopted strategies are summarized (section~\ref{sec:vlm_architectural}). Then, brief summary of models focusing on visual instruction tuning is provided (section~\ref{sec:v_tune}). Lastly, a brief overview is given for models that do not solve a direct pathology-specific task but perform other types of vision-language tasks (section~\ref{sec:vlm_related}).

\subsection{Categorization of VLMs}\label{sec:cat_vlm}
A categorization of VLMs in CPath can be done based on the reason for using the language modality. Some research works solve tasks like caption generation (\cite{zhang2020evaluating,tsuneki2022inference,qin2023whole,sengupta2023automatic,elbedwehy2024enhanced}) or VQA (\cite{naseem2022vision,chen2024wsi}) which necessitates utilization of both vision and language modality because of the nature of tasks. For VQA or caption generation, the generated output from the model needs to be in language form i.e. the model needs to perform a language task. On the other hand, other research works use language modality as a source of semantic information to be injected into information gained from vision modality. This additional semantic information can significantly boost the performance of the model which would not have been possible with a vision-only model. As an example, MI-Zero~\cite{lu2023visual} performs cancer subtyping which is not a language task but utilizes pathology reports curated from the in-house dataset and TCGA as a source of semantic information. 

\begin{figure}[!ht]
    \centering
    \includegraphics[width=0.7\linewidth]{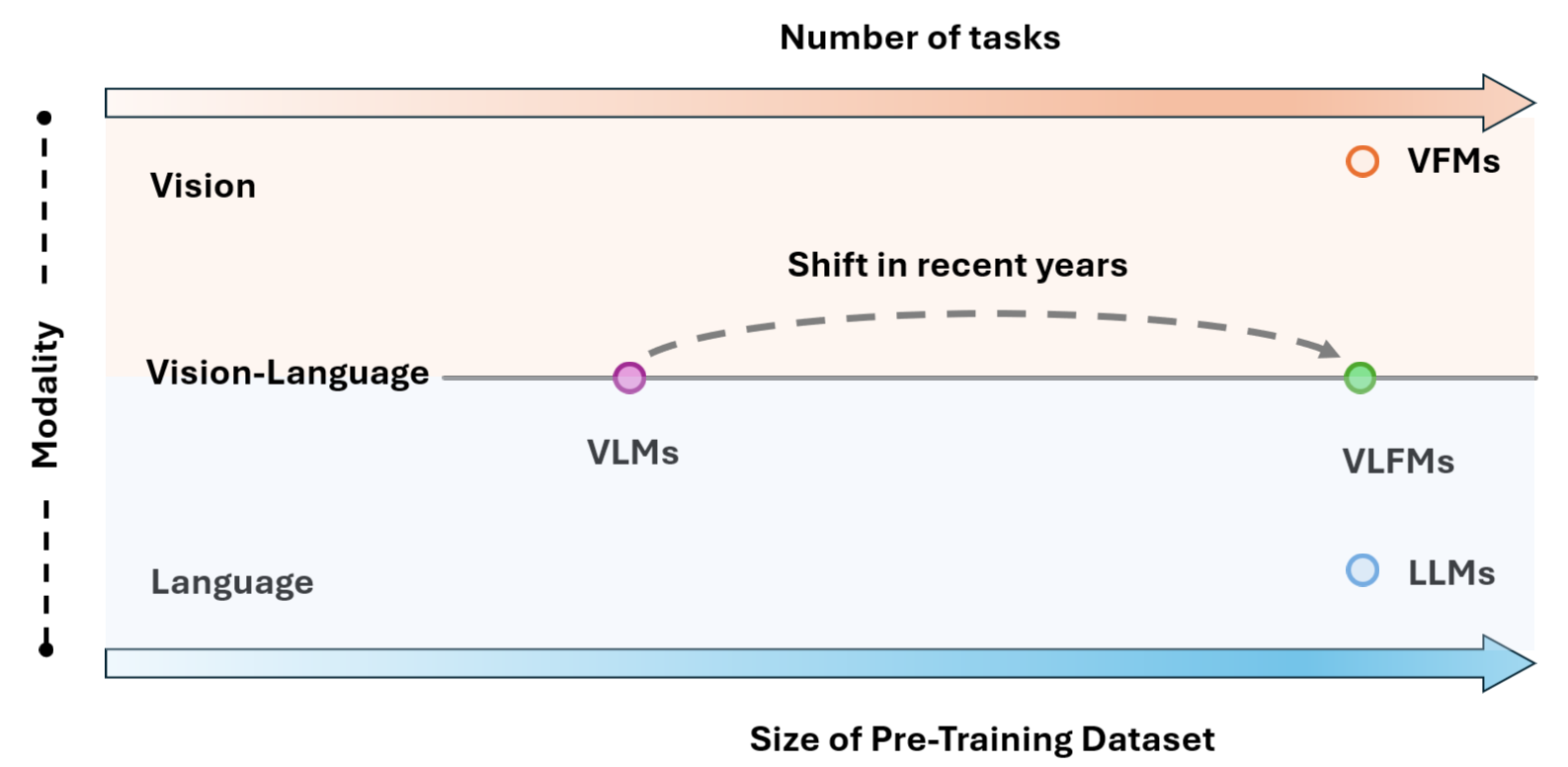}
    \caption{Visualization of categorization of VLMs and VLFMs. There are no clear distinctive criteria for a model to be classified as a VLFM or a VLM.}
    \label{fig:shift}
\end{figure}

Another categorization can be done by comparing the training approach with FMs.
VLFMs are different from VLMs as traditional VLMs focus on solving one or two vision-language tasks like caption generation. However, as FMs become more prevalent recent works are starting to shift towards VLFMs. Recent works (PathGen-CLIP~\cite{sun2024pathgen}, Quilt-Net~\cite{ikezogwo2024quilt}) take FM-like approach to train their models and adapt them to different downstream tasks. On the other end of the spectrum, there are VLMs that solve only a single task. As shown in Fig.~\ref{fig:shift} VLMs and VLFMs should be viewed as a spectrum where if the number of tasks and pre-training data size increases it moves away from VLMs and towards VLFMs.

\begin{table}[!ht]
\centering
\caption{Summary of pre-training strategies in VLMs}
\renewcommand{\arraystretch}{1.2}
\setlength{\tabcolsep}{4pt}
\resizebox{0.85\linewidth}{!}{
\begin{tabular}{@{}p{0.5\textwidth}p{0.15\textwidth}@{}}
\toprule
\textbf{Strategy}                               & \textbf{Reference} \\ \midrule
Performs pre-training from scratch (vision or language module)             & \makecell[l]{\cite{ikezogwo2024quilt},\cite{sun2024pathgen}}                 \\ \midrule
Utilize domain-specific vision module (HIPT, CTransPath,  etc) or language module (PubMedBERT, BioClinicalBERT, etc)  & \makecell[l]{\cite{zhang2024prompting},\cite{sengupta2023automatic},\cite{hu2024histopathology}, \\ \cite{guevara2023caption},\cite{chen2024wsi}, \cite{chen2023mi}, \\ \cite{tran2024generating}}                 \\ \midrule
Initialize with out-of-domain encoders (ImageNet pre-training, pre-trained CLIP, etc) & \makecell[l]{\cite{tsuneki2022inference}, \cite{qin2023whole}, \cite{elbedwehy2024enhanced}\\ \cite{watawana2024hierarchical}} \\  \bottomrule
\end{tabular}}
\label{tab:vlm_pretraining}
\end{table}

Owing to this, some VLMs follow the FM approach of pre-training the vision or text module. However, as pre-training from scratch requires a large amount of data and computing some VLMs use domain-specific vision modules from HIPT, CTransPath, IBMIL, etc or language modules from PubMedBERT, BioClinicalBERT, etc. This allows the vision or language module to learn an initial vision or language representation space which might not be as rich as pre-training scratch but boosts the performance. A summary of the choice of encoder type and pre-training strategy is provided in Table~\ref{tab:vlm_pretraining}.

\begin{figure}[!ht]
    \centering
    \includegraphics[width=0.7\linewidth]{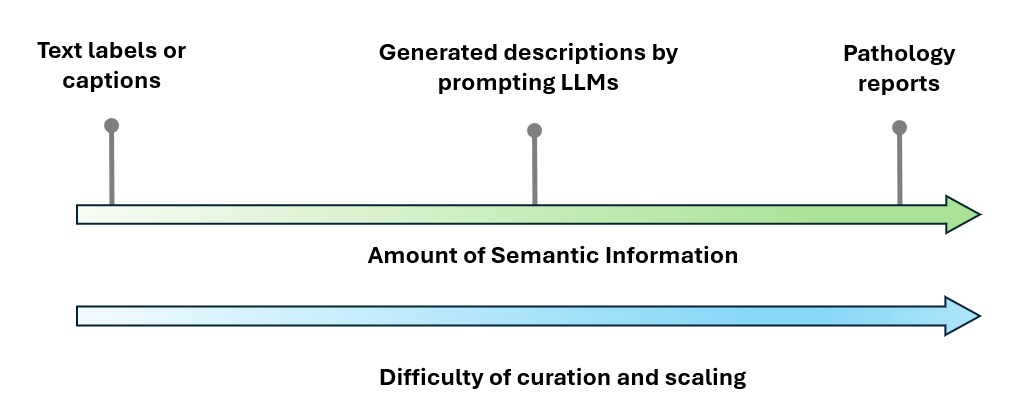}
    \caption{Amount of semantic information available through different text sources.}
    \label{fig:si}
\end{figure}

Another categorization can be done by analyzing how language prior is incorporated into the model. Most works use text labels or captions curated from different sources, which provide the least amount of information. Recent works like FSWC~\cite{qu2024rise}, ViLA-MIL~\cite{shi2024vila} and PEMP~\cite{qu2024pathology} utilize prompting LLMs to increase the amount of semantic information that can be gained from text labels or captions. They prompt the LLMs to generate descriptions about a particular class label or of morphological or textural patterns about patches. On the other end of the spectrum is research work utilizing all information from pathology reports which contain a lot of details. However, as shown in Fig.~\ref{fig:si} the difficulty of curation and doing it at a scale to match the pathologist level also increases. 

\subsection{Architectural components for VLMs in CPath}\label{sec:vlm_architectural}
\begin{figure}
    \centering
    \includegraphics[width=0.7\linewidth]{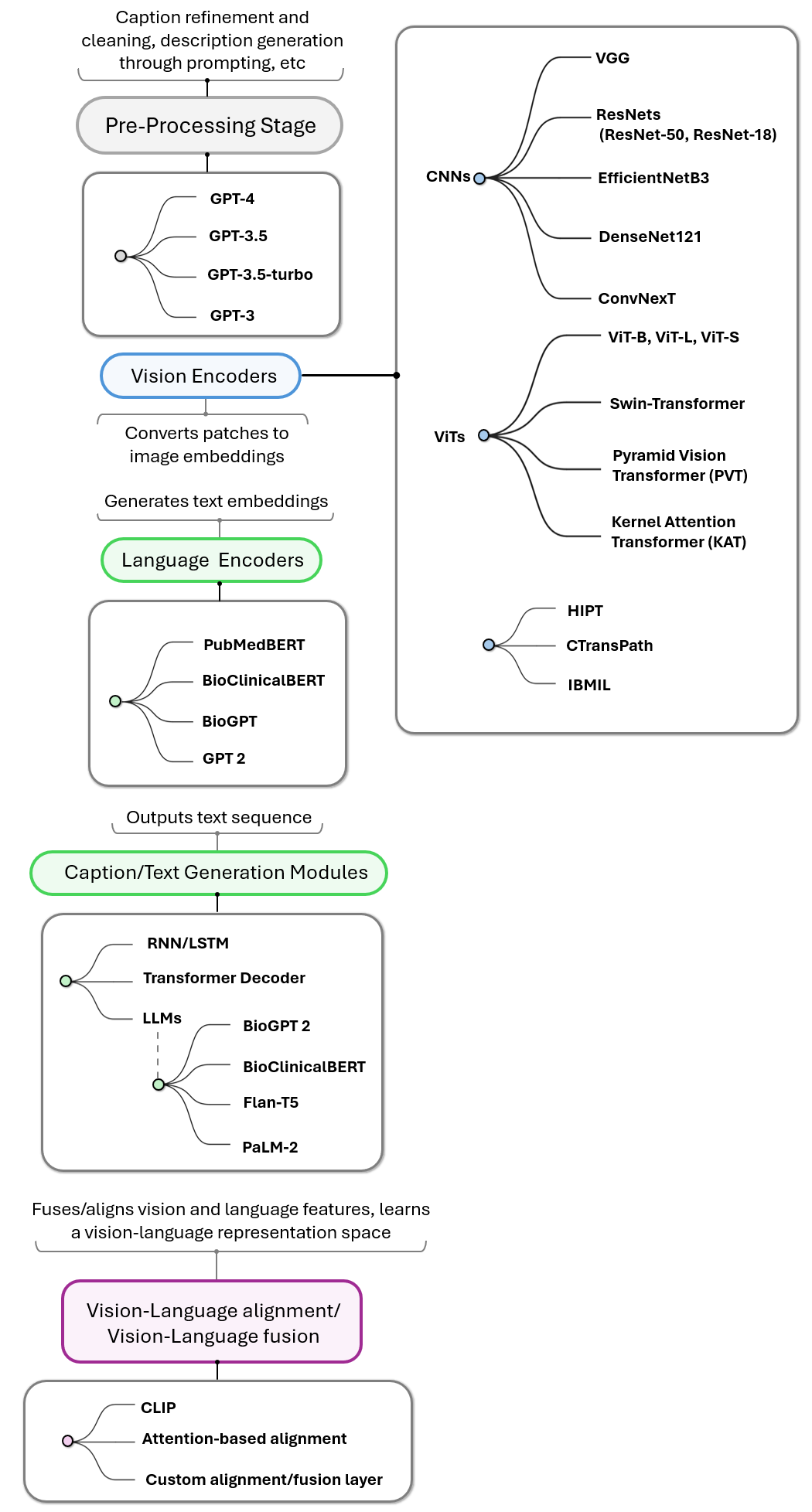}
    \caption{Common modules used in different stages of VLMs. The last stage is the vision-language alignment or fusion stage where the vision and language embeddings generated from the earlier stages are combined.}
    \label{fig:vlm_modules}
\end{figure}

Even though there is a huge variety of vision and language modules used in VLMs, it is possible to identify common structures. In Fig.~\ref{fig:vlm_modules} the common architectural components used in different stages are outlined. In the pre-processing stage, LLMs like GPT-4, GPT-3.5, GPT-3.5-turbo and GPT-3 are used to clean and refine captions or prompting them to generate descriptions. Examples of research work utilizing LLMs in the pre-processing phase include HistGen, FSWC, ViLA-MIL, FiVE, CPLIP and PromptBio. 

Another key component of the architecture is the vision module or encoder which converts the WSI patches to image embeddings. As shown in Fig.~\ref{fig:vlm_modules} it can be separated into three separate groups. The first group is vanilla CNNs that have ImageNet initialization or some kind of pre-training on pathology data. Note that most VLMs utilizing these architectures are not recent works (with the exception of Elbedwehy et al.~\cite{elbedwehy2024enhanced}). Recent VLMs utilize ViTs which is the second group containing several ViT variants. The third group is special encoders that are pathology domain-specific and optimized for encoding WSI patches. These are ideal for learning a rich representation without the cost of pre-training from scratch.  

For language encoders, most works utilize BERT (PubMedBERT, BioClincialBERT) or GPT (BioGPT, GPT 2) variants and most are pre-trained on datasets from the biomedical domain.

Another type of language module is the caption/text generation module which rather than encoding text into embedding, generates text sequences. Earlier works included RNN, LSTM and Transformer decoders and recent works include LLMs.

The most common vision-language alignment or fusion module is the CLIP model and its variations. However, some works come up with custom approaches to align vision and language representations.

\subsection{Visual Instruction Tuning in CPath}\label{sec:v_tune}

There are four research works/models that fall in this category (PA-LLaVA~\cite{dai2024pa}, PathInsight~\cite{wu2024pathinsight}, CLOVER~\cite{chen2024cost}, PathGen-LLaVA~\cite{sun2024pathgen} and Quilt-LLaVA~\cite{seyfioglu2024quilt}).

In PA-LLaVA, a domain-specific large language-vision assistant is proposed which is a LLaVA-based model. In terms of architecture, PLIP is used as the vision encoder and  LLama3 is used as the LLM (in conjunction with LoRA~\cite{hu2021lora}). A connector module that consists of self-attention and cross-attention blocks is leveraged to project image embeddings to match the dimension of language tokens. PathInsight leverages three different multi-modal models LLaVA, Qwen-VL~\cite{bai2023qwen} and InternLM~\cite{zhang2023internlm} (in conjunction with LoRA or full-parameter tuning). As the vision encoder either ViT or CLIP is used. CLOVER~\cite{chen2024cost} proposes a framework for cost-effective instruction learning in CPath. It utilizes Quilt-1M as the vision-language dataset, BLIP-2~\cite{li2023blip} for vision-language pre-training and a frozen visual encoder (EVA-ViT-G/14~\cite{fang2023eva}) and a frozen LLM (FlanT5~\cite{chung2024scaling} or Vicuna~\cite{vicuna2023}). The next work PathGen-LLaVA replaces the vision encoder for the CLIP model in LLaVA-v1.5 with PathGen-CLIP and as the frozen LLM Vicuna is used. The last work in this category Quilt-LLaVA is initialized with the general-domain LLaVA. Then it is tuned with QUILT for histopathological domain-alignment and finally, Quilt-instruct is used for visual instruction tuning.

\subsection{Related works for VLMs in CPath}\label{sec:vlm_related}
There are several research works that do not directly utilize VLMs to solve a task in the pathology domain, but investigate or use VLMs for other reasons. One example is Thota et al~\cite{thota2024demonstration} which investigates the effect of projected gradient descent (PGD)
adversarial perturbation attack on PLIP~\cite{huang2023visual} architecture. Another work by Lucassen et al.~\cite{lucassen2024preprocessing} provides a pathology report processing workflow that can be used for VLMs in CPath.

Another category of VLMs in CPath that does not directly solve pathology-specific tasks is text-guided diffusion models for image generation. However, these models can perform tasks like virtual stain transfer which can later be used to solve a pathology-specific task. Another use case is extending dataset size with generated synthetic images. One such work is PathLDM~\cite{yellapragada2024pathldm}, that performs text-to-image generation on the TCGA-BRCA dataset.

\section{Conclusion}\label{sec:conclusion}

In recent years the number of research works in CPath with FMs and VLMs has increased significantly which provides an indication about how CPath will evolve in the next couple of years. Many of these works put significant computing resources into training FMs on massive pre-training datasets or coming up with novel strategies to push more language prior knowledge into VLMs. In the near future, the VLFMs which bring together the benefits of both FMs and VLMs, are going to be the dominant model. This review article provides a comprehensive overview of all these models which will aid future researchers.

\bibliography{ref.bib}
\bibliographystyle{IEEEtran}

\end{document}